\documentclass[lettersize,journal]{IEEEtran}
\usepackage{amsmath,amsfonts}
\usepackage{algorithmic}
\usepackage{algorithm}
\usepackage{array}
\usepackage[caption=false,font=normalsize,labelfont=sf,textfont=sf]{subfig}
\usepackage{textcomp}
\usepackage{stfloats}
\usepackage{url}
\usepackage{verbatim}
\usepackage{graphicx}
\usepackage{cite}
\usepackage{times}
\usepackage{epsfig}
\usepackage{amsmath}
\usepackage{amssymb}
\usepackage{cite}
\usepackage{booktabs}
\usepackage{multirow}
\usepackage{bbm}
\usepackage{makecell}
\usepackage{comment}
\usepackage{color}

\newcommand{\etal}{\textit{et al.}}
\newcommand{\ptitle}[1]{\textbf{#1}\hspace{5pt}}

\newtheorem{theorem}{\bf Theorem}

\begin{document}

\title{SphereNet: Learning a Noise-Robust and General Descriptor for Point Cloud Registration}

\author{
Guiyu~Zhao, Zhentao Guo, Xin Wang, and Hongbin~Ma,~\IEEEmembership{Senior~Member,~IEEE}
\thanks{This work was partially funded by the National
Key Research and Development Plan of China (No.
2018AAA0101000) and the National Natural Science
Foundation of China under grant 62076028 (Corresponding author: Hongbin~Ma)}
\thanks{Guiyu~Zhao Zhentao Guo, Xin Wang, and Hongbin~Ma are with the National Key Lab of Autonomous Intelligent Unmanned Systems, School of Automation, Beijing Institute of Technology, 100081, Beijing, P. R. China
(e-mail: 3120220906@bit.edu.cn, zt\_guo1230@163.com, 18737173446@163.com, mathmhb@bit.edu.cn).}
}

\markboth{Journal of \LaTeX\ Class Files,~Vol.~14, No.~8, August~2021}%
{Shell \MakeLowercase{\textit{et al.}}: A Sample Article Using IEEEtran.cls for IEEE Journals}

\maketitle

\begin{abstract}
Point cloud registration is to estimate a transformation to align point clouds collected in different perspectives.
In learning-based point cloud registration, 
a robust descriptor is vital for high-accuracy registration. 
However, most methods are susceptible to noise and have poor 
generalization ability on unseen datasets. 
Motivated by this, we introduce SphereNet to 
learn a noise-robust and unseen-general descriptor 
for point cloud registration.
In our method, first, the spheroid generator builds a geometric domain 
based on spherical voxelization to encode initial features.
Then,
the spherical interpolation of the sphere is introduced to realize robustness against noise.
Finally, a new spherical convolutional 
neural network with spherical integrity padding completes the extraction of descriptors, 
which reduces the loss of features and fully captures the geometric features.
To evaluate our methods,  a new benchmark 3DMatch-noise with strong noise is introduced.
Extensive experiments are carried out on both indoor and outdoor datasets.
Under high-intensity noise, SphereNet increases the feature matching recall by more than 25 percentage points 
on 3DMatch-noise. In addition, it sets a new state-of-the-art performance for the 3DMatch and 3DLoMatch
benchmarks with 93.5\% and 75.6\% registration recall
and also has the best generalization ability on unseen datasets.
\end{abstract}

\begin{IEEEkeywords}

   Point cloud registration, feature learning, anti-noise ability, generalization ability.
   
   \end{IEEEkeywords}

\section{Introduction}

\begin{figure}[!t]
   \centering{\includegraphics[scale=0.36]{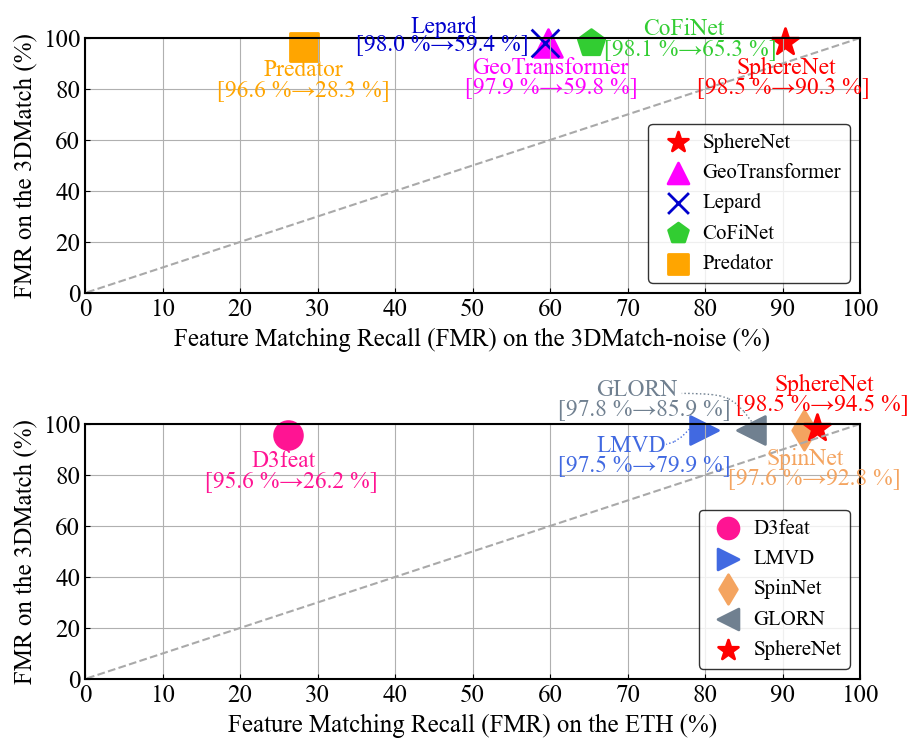}}  
   \caption{Feature Matching Recall on the 3DMatch \cite{zeng20173dmatch},
   ETH dataset \cite{pomerleau2012challenging}, and 3DMatch-noise benchmark.  
   Our method not only achieves the highest FMR on 3DMatch, 
   but also has the best generalization ability and anti-noise ability.}
   \label{FMR}
   \vspace{-10pt}
   \end{figure}

Consistently aligning 3D point clouds from different 
views to the same view is called 3D point cloud registration.
As a very important task in computer vision, 3D point cloud registration has 
a wide range of applications in medical science~\cite{hu2022cross}, robotics, and other fields.

Before deep learning was widely used, 
handcrafted descriptors has played a great role in point cloud registration.
In the early years, many handcrafted descriptors 
such as spin image (SI)\cite{Johnson1997SpinImagesAR}, point feature histogram (PFH)\cite{PFH}, fast point feature histogram
(FPFH)\cite{FPFH}, and signature of histogram of orientation (SHOT)\cite{salti2014shot} are proposed to try to extract
local features and geometric information. These traditional methods can extract 
geometric information and have better interpretability than deep learning. 
However, in the face of unseen scenes and noisy real scenes, the correct registration is 
difficult to achieve.

In recent years, deep learning\cite{voulodimos2018deep} has performed its dominance in various tasks in
computer vision. Thanks to deep learning and 3D datasets \cite{zeng20173dmatch, pomerleau2012challenging, geiger2012we}, point cloud registration\cite{huang2021comprehensive} has 
developed rapidly, and learning-based methods have flourished. 
However, the learning-based point cloud registration faces two limitations.
First, most of recent works \cite{huang2017coarse, yu2021cofinet,qin2022geometric,li2022lepard,yu2022riga, yangone} 
achieve high registration recall scores through a coarse-to-fine framework, but they cannot reject
outliers from a coarse scale which leads to false coarse correspondences and false point correspondences. And
the scenes with much noise will introduce many outliers.
Therefore, they are sensitive to noise and perform poorly in the scenes with large amounts of noise.
Second, although the approaches \cite{yu2021cofinet, qin2022geometric,huang2021predator} achieve excellent results 
on 3DMatch and KITTI, they perform poorly on unseen datasets with low generalization ability, which has some limitations
in the actual implementation. As a result, the descriptors made by 
\cite{yu2021cofinet,qin2022geometric,li2022lepard,yu2022riga,huang2021predator} 
are not robust and general and fails in most scenes with strong noise. 
In order to solve the adverse effects of noises and outliers on registration, 
some methods of iniler estimation~\cite{wu2022inenet,zhang2022vrnet} and outlier removal~\cite{wu2023sacf,chen2022sc2} are proposed. 
They have greatly improved the accuracy of registration, 
but most of them are limited to the correspondence 
filtering, and they do not fundamentally solve the problem in features.
In the scenes with strong noise, when the extracted descriptor is affected, 
these methods~\cite{wu2022inenet,zhang2022vrnet,wu2023sacf,chen2022sc2,yang2021toward} 
can only play a very limited role.


The performance of recent works~\cite{ao2021spinnet, yu2021cofinet,qin2022geometric} on 3DMatch 
is close to reaching saturation, 
but most works perform poorly in the actual implementation, due to two limitations described above. 
Therefore, this paper mainly focuses on the anti-noise ability and 
generalization ability. 
On this basis, 3DMatch-noise benchmark is proposed to evaluate the anti-noise ability of our method.
For this motivation, this paper introduces a new feature extraction method 
that can learn a noise-robust and unseen-general descriptor for 
registration in noisy and unseen scenes.

Our method is named SphereNet and consists of two modules. 
The first is the spheroid generator. It generates a voxel sphere 
to fully encode the initial geometric feature. In addition, given
some ingenious improvements, the module also achieves rotation invariance 
and anti-noise robustness. The other module is the spherical feature extractor. 
In this module, we introduce a new spherical convolutional neural network
with spherical integrity padding, 
which fully extracts noise-robust and unseen-general descriptors.
Our SphereNet has four key properties: 
(1) It realizes good anti-noise robustness. In the face of scenes with strong noise, 
it can also perform well.  
(2) The descriptor is descriptive and distinctive. 
(3) It realizes translation and rotation invariance. 
(4) It has excellent generalization ability on unseen datasets,
which is important for implementation in different real-world scenes.

In extensive experiments, our approach achieves the best performance 
compared with state-of-the-arts \cite{ao2021spinnet,yu2021cofinet,qin2022geometric}
in both indoor and outdoor scenes, achieving 93.4\% and 75.6\% 
registration recall on the 3DMatch and 3DLoMatch, respectively.
As shown in Figure \ref{FMR}, 
compared with state-of-the-arts~\cite{ao2021spinnet,qin2022geometric,li2022lepard,yu2021cofinet,huang2021predator},
the critical advantage of our method are 
excellent anti-noise robustness and generalization ability.
In the scenes with high-intensity noises, 
our SphereNet improves
the feature matching recall by 25.0 percentage points~(pp)
and the registration recall by 22.2 pp on the challenging
3DMatch-noise benchmark: Noise 1.
It also has the best generalization ability 
with FMR scores of 94.5\% and 87.2\% on the ETH and KITTI datasets.
The main contributions of this paper are as follows:
\begin{itemize}
\setlength{\itemsep}{0pt}
\setlength{\parsep}{0pt}
\setlength{\parskip}{0pt}
\item[$\bullet$] We introduce a new method for 3D point cloud registration
that can extract a rotation-invariant, noise-robust, and unseen-general descriptor. 
\item[$\bullet$] A new benchmark, 3DMatch-noise, is introduced to evaluate 
the anti-noise robustness of our method. 
In addition, our method achieves the best anti-noise ability 
compared with state-of-the-art methods.
\item[$\bullet$] A new feature encoding idea is proposed. The 
spherical voxelization with spherical interpolation 
fully encodes the geometric feature, which makes our method more robust to noise.
\item[$\bullet$] A new spherical convolutional neural network is introduced. With spherical integrity padding,
it reduces the loss of geometric features and learns a robust and general descriptor.
\end{itemize}

\section{Related Work}

\subsection{3D Handcrafted Descriptors}
In the early years, 3D handcrafted descriptors 
are used to complete the registration based on feature matching.
According to the region of the descriptor, it can be divided 
into global features and local features. 
The global features describe the whole point cloud, 
including viewpoint feature histogram (VFH)\cite{VFH} and clustered viewpoint 
feature histogram (CVFH)\cite{CVFH}. 
Local features\cite{guo2016comprehensive} focus on the local region at the patch level
which is better able to describe the local details. 
Point feature histogram (PFH)\cite{PFH} and fast point 
feature histogram (FPFH)\cite{FPFH} describe the local 
geometric features through the quaternion
information of angles and distance. 
Point pair feature (PPF)\cite{PPF}  
extracts features by calculating the normals and relative positions between point pairs. Tombari \etal propose 
signatures of histograms of orientations (SHOT)\cite{salti2014shot} based on the point signature (PS)\cite{chua1997PS}, 
which attempts to fully describe geometric information by constructing spherical domains.
With the help of the local reference frame (LRF), SHOT preliminarily realizes rotation invariance,
but it is still susceptible to noise.
Later, some other manual descriptors \cite{zhao2022method, guo2013rotational, GASD} based on LRF are proposed. 
Although they achieve good results in specific scenes, 
they are less effective in real-world scenes and sensitive to noise.


\subsection{3D Learning-based Descriptors}

Learning-based registration methods 
extract descriptors through feature extraction networks, 
and then use the robust descriptors to register point clouds. 
Compared with handcrafted descriptors 
\cite{FPFH,salti2014shot,PPF,guo2013rotational},  recent
learning-based methods \cite{huang2021predator,ao2021spinnet, qin2022geometric,huang2021predator,yu2021cofinet,wang2021you} 
have a stronger discriminative ability and generalization ability.
The learning-based descriptors can be divided into two categories 
\cite{ao2022YOTO}: patch-based descriptors
and fragment-based descriptors.

\ptitle{Patch-based Descriptor.}
The input of the patch-based descriptor is the local patch in the scene, 
while the input of the fragment-based descriptor is the whole fragment.
3DMatch~\cite{zeng20173dmatch} is the pioneering work in patch-based descriptors, which extracts descriptors in  the local patches by 3D 
convolutional neural networks (CNNs).
PPFNet~\cite{deng2018ppfnet} combines point pair features with pointnet \cite{qi2017pointnet} and is trained by 
N-tuple loss to complete global matching using local features. 
Then, PPF-FoldNet\cite{deng2018ppf} improves the invariance against rotation from PPFNet. 
SpinNet\cite{ao2022YOTO,ao2021spinnet} extracts features by cylindrical 
convolution, which achieves state-of-the-art performance on 
generalization ability.

\ptitle{Fragment-based Descriptor.}
Taking the entire fragment as input, FCGF~\cite{choy2019fully} uses the Minkowski CNNs \cite{choy20194d} 
and the U network \cite{ronneberger2015u, he2016deep} to extract a dense feature descriptor.
Predator \cite{huang2021predator} proposes an overlapping region attention module and 
uses the GNN to extract the local feature, which achieves a great 
improvement on the 3DLoMatch benchmark and provides a good 
solution in the scenes with low overlap.
GeoTransformer \cite{qin2022geometric} uses the Transformer~\cite{vaswani2017attention} and superpoint-to-point 
framework to  learn geometric features, which achieves state-of-the-art performance 
on 3DLoMatch. 

Overall, the methods of fragment-based descriptors are 
much faster than the methods of patch-based descriptors,
while the methods of patch-based descriptors can
capture local features in detail 
and have good generalization ability.
Unfortunately, both of them are sensitive to noise, 
and there is not a good way to deal with scenes with strong noise.



\section{SphereNet}
\begin{figure*}[!t]
\centering{\includegraphics[scale=0.61]{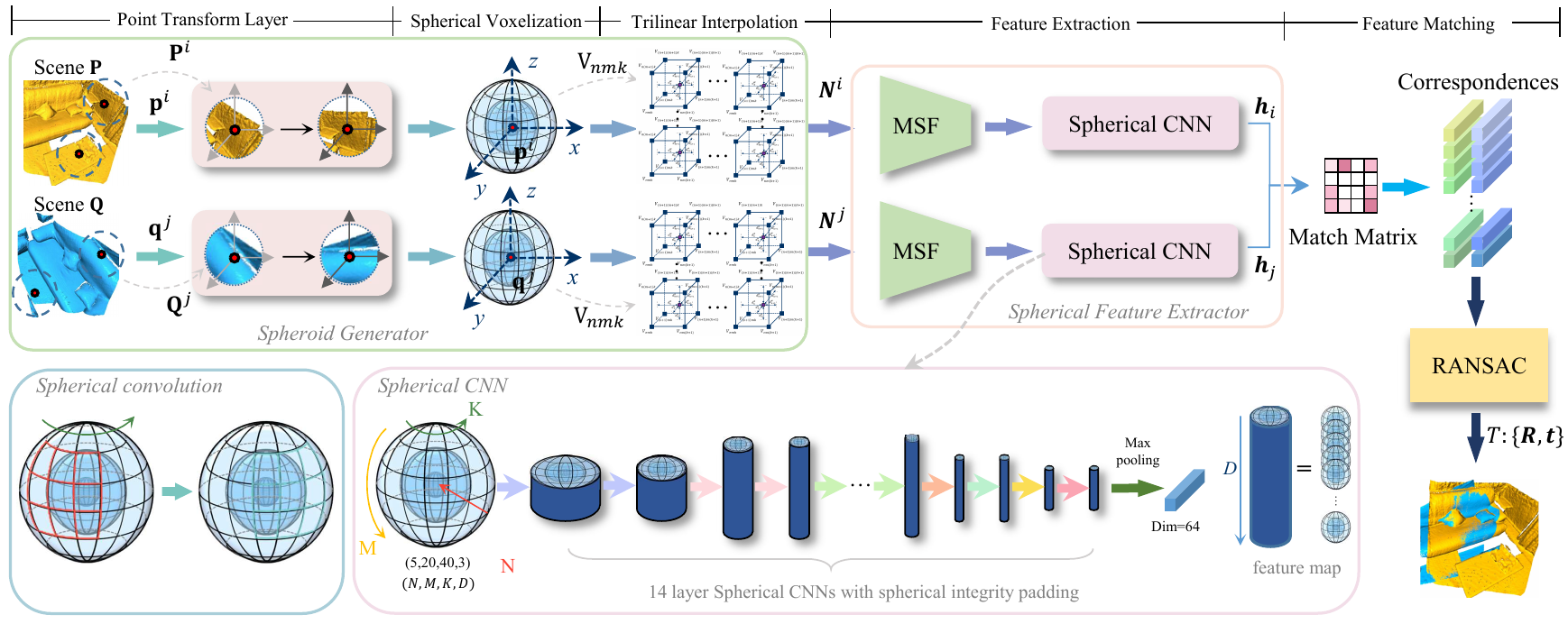}}  
\caption{
   Overall framework of our method.
In the spheroid generator, first, the superpoints $\mathbf{p}^i$ and $\mathbf{q}^j$ are downsampled from the fragments of source $\mathbf{P}$ and target $\mathbf{Q}$. 
The patches $\mathbf{P}^i$ and $\mathbf{Q}^j$ are obtained by searching for nearest neighbors at $\mathbf{p}^i$ and $\mathbf{q}^j$.
Second, the point transform layer is used to construct repeatable LRFs 
for patches $\mathbf{P}^i$ and $\mathbf{Q}^j$ to realize rotation invariance.
In order to improve the ability of noise resistance of the feature, 
spherical voxelization with spherical interpolation is carried out to encode the initial geometric features $\boldsymbol{N}^i$ and $\boldsymbol{N}^j$. 
In the spherical feature extractor, first, by multiscale  feature fusion, the generalization
ability and anti-density invariance of our SphereNet are improved.
After that, spherical CNN with spherical integrity padding is used to extract the robust and general descriptors $\boldsymbol{h_i}$ and $\boldsymbol{h_j}$. 
Finally, RANSAC is used to realize global matching to complete the registration.
}
\label{fig1}
\end{figure*}

\subsection{Problem Statement}
For a large-scale scene, two point clouds with overlapping areas
are generated from different perspectives, one point cloud is named source  
$\mathbf{P}=\left\{\mathbf{p}_i \in \mathbb{R}^3 \mid i=1, \ldots, I\right\}$
and another point cloud 
$\mathbf{Q}=\left\{\mathbf{q}_j \in \mathbb{R}^3 \mid j=1, \ldots, J\right\}$
is named target. The task of registration is to find a rigid transform 
$\mathbf T\{\mathbf R,\mathbf t\}$ where $\mathbf{R} \in SO(3)$ and 
$\mathbf{t} \in \mathbb{R}^3$, which aligns the 
two point clouds $\mathbf{P}$ and $\mathbf{Q}$.
The source is rotated and translated through the rigid transformation $\mathbf T$ to obtain
the transformed source $\mathbf T(\mathbf P)$, which 
minimizes the Euclidean distance between the correspondences
in the overlapping area of the $\mathbf T(\mathbf P)$ and $\mathbf Q$:
\begin{equation}
   \underset{\mathbf{R}, \mathbf{t}}{\arg \min } \sum_{j=1}^J \sum_{i=1}^I m_{ij} (\left\|\mathbf{R} \cdot \mathbf{p}_i+\mathbf{t}-\mathbf{q}_j\right\|_2^2)
   \end{equation}
where the value of $m_{ij}$ is 0 or 1, which is calculated by
\begin{equation}
m_{i j}= \begin{cases}1 & \text { if } (\mathbf{p}_i,\mathbf{q}_j) \in \mathcal{G} \\ 0 & \text { otherwise }\end{cases}
\label{mij}
\end{equation}
where $\mathcal{G}$ is the set of correspondences between $\mathbf{P}$ and $\mathbf{Q}$. Since the 
correspondences are unknown, so we should calculate the correspondences by feature matching.

As shown in Figure \ref{fig1}, our SphereNet is a learning-based registration method which will find the 
correspondences of point pairs with similar features.
Through our SphereNet, the feature $\boldsymbol{h}$ is extracted.
Then, the point clouds with similar
features $\boldsymbol{h}_{\mathbf{p}_i}$ and $\boldsymbol{h}_{\mathbf{q}_j}$ form a correspondence.
Finally, according to the correspondences, the rigid transformation $\mathbf T\{\mathbf R,\mathbf t\}$ is calculated  by RANSAC.

\subsection{Spheroid Generator}\label{Spheroid}
In order to facilitate the spherical convolutional neural network to fully extract
local geometric features, the spheroid generator is introduced. 
In the spheroid generator, we transform the point cloud of each patch from 
the global reference frame (GRF) to their local reference frames (LRFs), 
and disambiguate the direction of the LRFs according to the distribution of point clouds, 
which greatly improves the repeatability of LRFs. 
Then, under LRFs, the point cloud in a patch is spherically voxelized, which
encodes geometric features. 
Finally, the robustness of the spherical voxels against noise is improved 
by spherical interpolation.

\ptitle{Point Transform Layer.}\label{Point Transform Layer}
First, a superpoint $\mathbf{p}_i \in \mathbf{P}$ in a local region is obtained by 
random downsampling. 
By radius search, we search points in sphere $\mathbf{U}_{\mathbf{p}_i}^R$
to obtain the neighbors 
$\mathbf{P}^i=\{\mathbf{p}_{i_j}:\left\|\mathbf{p}_{i_j}-\mathbf{p}_i\right\|^2 \leq R\}$
where $R$ is the radius threshold and $\mathbf{p}_{i_j}$ is the $j$th neighbor of 
$\mathbf{p}_i$.
Then, we use $\mathbf{p}_i$ as the reference point to translate $\mathbf{p}_{i_j}$ to
$\widehat{\mathbf{p}_{i_j}}=\mathbf{p}_{i_j}-\mathbf{p}_i$.
LRFs~\cite{salti2014shot} are constructed by decomposing the point covariance matrix $\mathbf{M}$ in 
$\mathbf{U}_{\mathbf{p}_i}^R$ to achieve rotation invariance:
\begin{equation}
   \mathbf{M}=\sum_{\mathbf{p}_{i_j} \in \mathbf{P}^i} \frac{\left(R-d_{i_j}\right)}{\sum_{\mathbf{p}_{i_j} \in \mathbf{P}^i}\left(R-d_{i_j}\right)} \left(\mathbf{p}_{i_j}-\mathbf{p}_i\right)\left(\mathbf{p}_{i_j}-\mathbf{p}_i\right)^T
\label{M}
\end{equation}
where $d_{i_j}$ is the distance between the center point $\mathbf{p}_i$ 
and its $j$th neighbor $\mathbf{p}_{i_j}$.
Then, the LRFs are constructed by SVD.

\begin{theorem} \label{lemma1}
   \emph{
   Assume for any rotation transformation $R_{xyz} \in SO(3)$, and
   any XY-plane rotation transformation $R_{xy|z_n} \in SO(2)$ 
   fixed in the direction $z_n$. And the transformation $R_{z\rightarrow z_{n}} \in SO(3)$ 
   is obtained through the covariance matrix decomposition,
   which aligns the point cloud from direction $\mathbf{z}$-axis to $z_n$.
   The linear transformation $\mathcal{L} (\cdot)$ of the point transform layer 
   maps the local patch $\mathbf{P}^i$ to $\widehat{\mathbf{P}^i} : \mathbb{R}^{3 \times\left|\mathbf{P}^{i}\right|} \rightarrow \mathbb{R}^{3 \times\left|\mathbf{P}^{i}\right|}$,
   then there is an equivariant relation 
   $\mathcal{L} (R_{xyz} \circ \mathbf{p}^i)=R_{xy|z_n}\circ \mathbf{p}^i$.}
   \end{theorem}
   \begin{IEEEproof}[Proof of Theorem \ref{lemma1}]
   \emph{
   According to  the principle of rotation invariance, 
   the matrix of linear transformation $\mathcal{L}(\mathbf{P}^i)$ is expressed as 
   $R^{-1}_{z\rightarrow z_n}\circ \mathbf{P}^i=R_{z_{n}\rightarrow z}\circ \mathbf{P}^i$, 
   where $\mathbf z$ is the direction vector of the z-axis of the GRFs 
   and $z_{n}$ is the $\mathbf{z}$ direction vector of the z-axis of LRFs. Thus, we have:
   }
   \begin{equation}
      \begin{aligned}
      \mathcal{L} (R_{xyz} \circ \mathbf{P}^i)
      &=R^{-1}_{z_\rightarrow z_{n}} \circ (R_{xyz} \circ \mathbf{P}^i) \\
      &=R_{z_{n}\rightarrow z} \circ (R_{xyz} \circ\mathbf{P}^i) \\
      &=(R_{z_{n}\rightarrow z} \cdot R_{xyz}) \circ \mathbf{P}^i \\
      &=R_{xy|z_n} \circ \mathbf{P}^i
      \end{aligned}
   \end{equation}

\end{IEEEproof}

\ptitle{Spherical Voxelization.}\label{Spherical voxelization}
Inspired by \cite{frome2004recognizing,lei2019octree,salti2014shot}, the sphere domain $\mathbf{U}_{\mathbf{p}_i}^R$ is divided
into $N\times M\times K$ small voxels $\mathbf{V}_{n m k}$ from three directions: 
radius length $r=\sqrt{x^2+y^2+z^2}$, elevation angle $\theta=\arccos\frac{z}{\sqrt{x^2+y^2+z^2}}$ and azimuth 
angle $\varphi=\arctan \frac{y}{x}$,
where $n \in\{0, \ldots, N\}$ 
$m \in\{0, \ldots, M\}$ and $k \in\{0, \ldots, K\}$ are indexes 
of $r $, $\theta $ and $\varphi $, respectively.
The voxel boundary sets $S_r$, $S_{\theta}$, and $S_\varphi$
of radius length $r $, 
elevation angle $\theta  $, and azimuth 
angle $\varphi $ are defined as follows:

\begin{equation}
\begin{cases}
S_r&=\{r_n: r_0+n\cdot \frac{r_N-r_0}{N} \}\\
S_{\theta}&=\{\theta_m: \theta_0+m\cdot \frac{\theta_M-\theta_0}{M}\}\\
S_\varphi&=\{\varphi_k: \varphi_0+k\cdot \frac{\varphi_K-\varphi_0}{K} \}
\end{cases}
\label{1}
\end{equation}
where $r_0=0$, $\theta_0=0$ and $\varphi_0=0$ are the first items of 
the three boundary sets and $r_N=R$, $\theta_M=\pi $ and $\varphi_K=2\pi$ are the last items.
By the boundary value sets $S_r$, $S_{\theta}$, $S_\varphi$, 
the sphere domain $\mathbf{U}_{\mathbf{p}_i}^R$ is voxelized to small voxels $\mathbf{V}_{n m k}$
and the points $\mathbf{P}^i_{n m k}$ 
in the voxel $\mathbf{V}_{n m k} \in \mathbf{U}_{\mathbf{p}_i}^R$ 
is defined as follows:
\begin{equation}
\{\mathbf{p}_{i_j}: \!(r_{n\!-\!1},\theta_{m\!-\!1},\varphi_{k\!-\!1})\!\leqslant \!(r^i_j,\theta^i_j,\varphi^i_j)\!<\!(r_n,\theta_m,\varphi_k)
\}
\end{equation}
where $(a_1,b_1,c_1)<(a_2,b_2,c_2)$ is equal to $(a_1\!<\!a_2) \land (b_1\!<\!b_2)\land (c_1\!<\!c_2)$.
Finally, the number $N_{n m k}=|\mathbf{P}^i_{n m k}|$
in the voxel $\mathbf{V}_{n m k}$ is counted as our initial description of geometric information.


\begin{figure*}[thb]
   \centering{\includegraphics[scale=0.48]{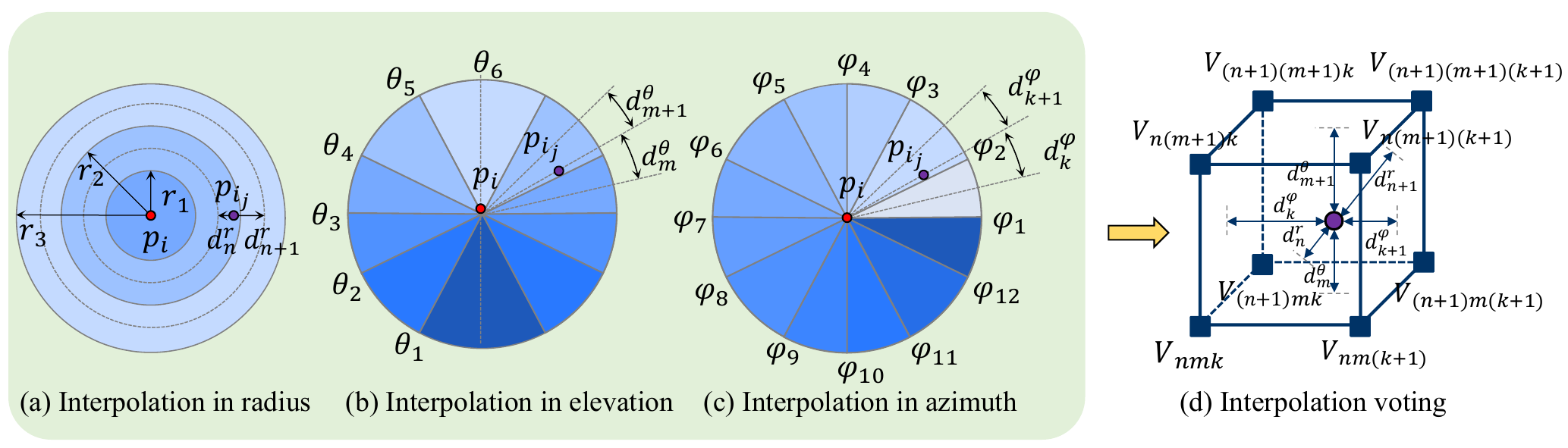}}  
   \caption{Illustration of spherical interpolation based on spherical voxelization.}
   \label{trilinear}
   \end{figure*}

\ptitle{Spherical Interpolation.}
Theoretically, the greater number of spherical voxels, 
the finer feature we obtain. However, considering the 
universality of features and the real-time performance of the algorithm, 
we have to reduce the number of spherical voxels. 
In addition, due to the large amount of noise in the actual 
scenes, 
many points $\mathbf{p}_{i_j} \in \mathbf{V}_{n_1 m_1 k_1}$ affected by noise $\boldsymbol\varepsilon$ may be allocated to another specific 
small voxel $\mathbf{p}_{i_j}+\boldsymbol\varepsilon \in \mathbf{V}_{n_2 m_2 k_2}$  
which finally causes the error of our initial feature.
Therefore, in order to overcome the influence of noise 
and extract finer features under a few voxels, 
this paper proposes a spherical interpolation method.
By this method, each point in each voxel is 
interpolated in the three dimensions of radius length $r $, 
elevation angle $\theta  $ and azimuth 
angle $\varphi $ respectively. 
Illustration of spherical interpolation 
based on spherical voxelization is shown in Figure \ref{trilinear}.

Interpolation in radius length. 
As shown in Figure \ref{trilinear} (a), 
the $N+1$ boundary values in the boundary set $S_r$ 
divide the whole sphere into $N$ regions in the dimension of radius length. 
In each region $n$, the region centerline (the dotted lines in Figure \ref{trilinear}) can be calculated as $\frac{(r_n+r_{n-1})}{2} $. 
Then, the interpolation weight $w^r_n$ from point $\mathbf{p}_{i_j}$ to region $n$
is given by the following equation:
\begin{equation}
   w^r_n=
   \begin{cases}
   1-\frac{d^r_n}{r_N/N}  &,  \quad  d^r_n < \frac{r_N}{N} \wedge r^i_j \geqslant \frac{r_N}{2N}\\
   1  &,  \quad  d^r_n < \frac{r_N}{N} \wedge r^i_j < \frac{r_N}{2N}\\
   0                                &,  \quad d^r_n \geqslant  \frac{r_N}{N}
   \end{cases}
   \end{equation}
where $d^r_n=|\frac{(r_n+r_{n-1})}{2}-r^i_j|$ is the distance
between point $\mathbf{p}_{i_j}$ and the region centerline of region $n$,
and $n \in\{1, \ldots, N\}$.

Interpolation in elevation angle. 
As shown in Figure \ref{trilinear} (b),
the $M+1$ boundary values in the boundary set $S_\theta$ 
divide the whole sphere into $M$ regions in the dimension of elevation angle $\theta$. 
In each region $m$, the region centerline can be calculated as $\frac{(\theta_m+\theta_{m-1})}{2} $. 
Therefore, the interpolation weight $w^\theta_m$ from point $\mathbf{p}_{i_j}$ to region $m$
is given by the following equation:
\begin{equation}
   w^\theta_m=
   \begin{cases}
   1-\frac{d^\theta_m}{{\theta_M}/{M}}  \!&,\!\quad  d^\theta_m < \frac{\theta_M}{M} \wedge \theta_M - \frac{\theta_M}{2M} \geqslant \theta^i_j \geqslant \frac{\theta_M}{2M}\\
   1  \!&,\!  \quad  d^\theta_m < \frac{\theta_M}{M} \wedge \theta^i_j < \frac{\theta_M}{2M}\\
   1  \!&,\!  \quad  d^\theta_m < \frac{\theta_M}{M} \wedge \theta^i_j > \theta_M - \frac{\theta_M}{2M}\\
   0                                \!&,\! \quad  d^\theta_m \geqslant \frac{\theta_M}{M}
   \end{cases}
   \end{equation}
where $d^r_n=|\frac{(\theta_m+\theta_{m-1})}{2}-\theta^i_j|$ is the distance between point $\mathbf{p}_{i_j}$ 
and the region centerline of region $m$, and $m \in\{1, \ldots, M\}$.

Interpolation in azimuth angle. 
The interpolation weight $w^\varphi_k$ from point $\mathbf{p}_{i_j}$ to region $k$
is given by the following equation:
\begin{equation}
   w^\varphi_k=
   \begin{cases}
   1-\frac{d^\varphi_k}{\varphi_K/K}  &,  \quad  d^\varphi_k < \frac{\varphi_K}{K} \\
   0  &,  \quad  d^\varphi_k \geqslant \frac{\varphi_K}{K}
   \end{cases}
   \end{equation}
where $d^\varphi_k=|\frac{(\varphi_k+\varphi_{k-1})}{2}-\varphi^i_j|$ is the distance between point $\mathbf{p}_{i_j}$ 
and the region centerline of region $k$, and $k \in\{1, \ldots, K\}$.

Then, according to the weights $w^r_n$, $w^\theta_m$, and $w^\varphi_k$
in three dimensions, the number $N^j_{n m k} $ that the $j$th neighbor $\mathbf{p}_{i_j}$ 
interpolated into the small voxel $\mathbf{V}_{n m k}$, 
can be calculated: 
\begin{equation}
N^j_{n m k}=w^r_n \times w^\theta_m \times w^\varphi_k
\label{N}
\end{equation}
As shown in Figure \ref{trilinear} (d), each point casts interpolation votes 
to the $2\times 2 \times 2$ nearest voxels (the eight vertices of the voxel).
Finally, the initial geometric features are encoded by 
summing the number of votes for each small voxel:
\begin{equation}
N_{n m k}=\sum_{\mathbf{p}_{i_j} \in \mathbf{P}_i} N^j_{n m k}  
\label{sumN}
\end{equation}

\begin{theorem}\label{lemma2}
   \emph{
   Given any XOY-plane rotation transformation $R_{xy|z_n} \in SO(2)$ 
   fixed in the direction $\mathbf{z}_n$ and 
   defined the neighbors $\mathbf{P}^i$ in the voxel $\mathbf{V}^i_{n m k}$ as 
   $\mathbf{P}^i_{nmk} \subseteq \mathbf{P}^i$, then the spherical voxelization 
   with spherical interpolation is an equivariant mapping $\mathcal{H} (\cdot)$ of the transformation $R_{xy|z_n}$.
   }
   \end{theorem}
   
   \begin{IEEEproof} [Proof of Theorem \ref{lemma2}]
   \emph{
   Let 
   $\mathcal{H} (\cdot)$ be a mapping from $\mathbf{P}^i$ 
   after the point transform layer and direction disambiguation to the initial feature 
   $\mathbf{N}^i=\{\mathbf{N}^i_{1 1 1},\ldots,\mathbf{N}^i_{n m k},\ldots,\mathbf{N}^i_{N M K}\}$ 
   in spherical voxel $\mathbf{V}^i$: 
   $\mathbb{R}^{3 \times\left|\mathbf{P}^{i}\right|} \rightarrow \mathbb{R}^{N \times M \times K}$.
   Since the rotation transformation $R_{xy|z_n}$ is only in the XOY-plane, 
   the points $\mathbf{P}^i_{nmk}$ located in small voxels 
   will be rotated to $\mathbf{P}^i_{nm(k+\zeta )}$ 
   with the same radius length $r$, elevation angle $\theta $ and different azimuth angle $\varphi  $, 
   where $\zeta$ is related to the rotation angle in the XOY-plane. 
   Further derivation, we can obtain: 
   }
   \begin{equation}
   \begin{aligned}
   \mathbf{R}_{xy|z_n}  \circ \mathbf{P}^i_{nmk}=\mathbf{P}^i_{nm(k+\zeta )} \\
   \mathbf{R}_{xy|z_n}  \cdot \boldsymbol{N}^i_{nmk}=\boldsymbol{N}^i_{nm(k+\zeta )}
   \end{aligned}
   \label{PN}
   \end{equation}
   \emph{
   where $\boldsymbol{N}^i_{nmk} $ is the number of points located in the voxel $\mathbf{V}^i_{nmk}$ after interpolation
   and it has the same position information as $\mathbf{V}^i_{nmk}$. 
   Therefore, $\mathbf{R}_{xy|z_n} $ is the transformation of the position in $\mathbf{V}^i_{nmk}$, 
   and $\boldsymbol{N}^i_{nm(k+\zeta )}$ is the number of points under the new voxel $\mathbf{V}^i_{nm(k+\zeta )}$.
   According to Eq.~\ref{PN}, we have completed the derivation of Lemma 2:
   }
   \begin{equation}
   \begin{aligned}
   &\mathbf{R}_{xy|z_n} \circ \mathcal{H}\left(\mathbf{P}^{i}\right)=\mathbf{R}_{xy|z_n} \circ \boldsymbol{N}^i\\
   &=\mathbf{R}_{xy|z_n} \circ\left[\boldsymbol{N}^{i}_{111}, \ldots, \boldsymbol{N}^{i}_{n m k}, \ldots, \boldsymbol{N}^{i}_{N M K}\right] \\
   &=\left[\boldsymbol{N}^{i}_{11(1+\zeta )}, \ldots, \boldsymbol{N}^{i}_{n m (k+\zeta )}, \ldots, \boldsymbol{N}^{i}_{N M (K+\zeta )}\right] \\
   &=\mathcal{H}\left(\mathbf{P}^i_{nm(k+\zeta )}\right)=\mathcal{H}\left(\mathbf{R}_{xy|z_n} \circ \mathbf{P}^{i}\right)
   \end{aligned}
   \end{equation}
   \end{IEEEproof}

\begin{figure}[htbp]
   \centering{\includegraphics[scale=0.2]{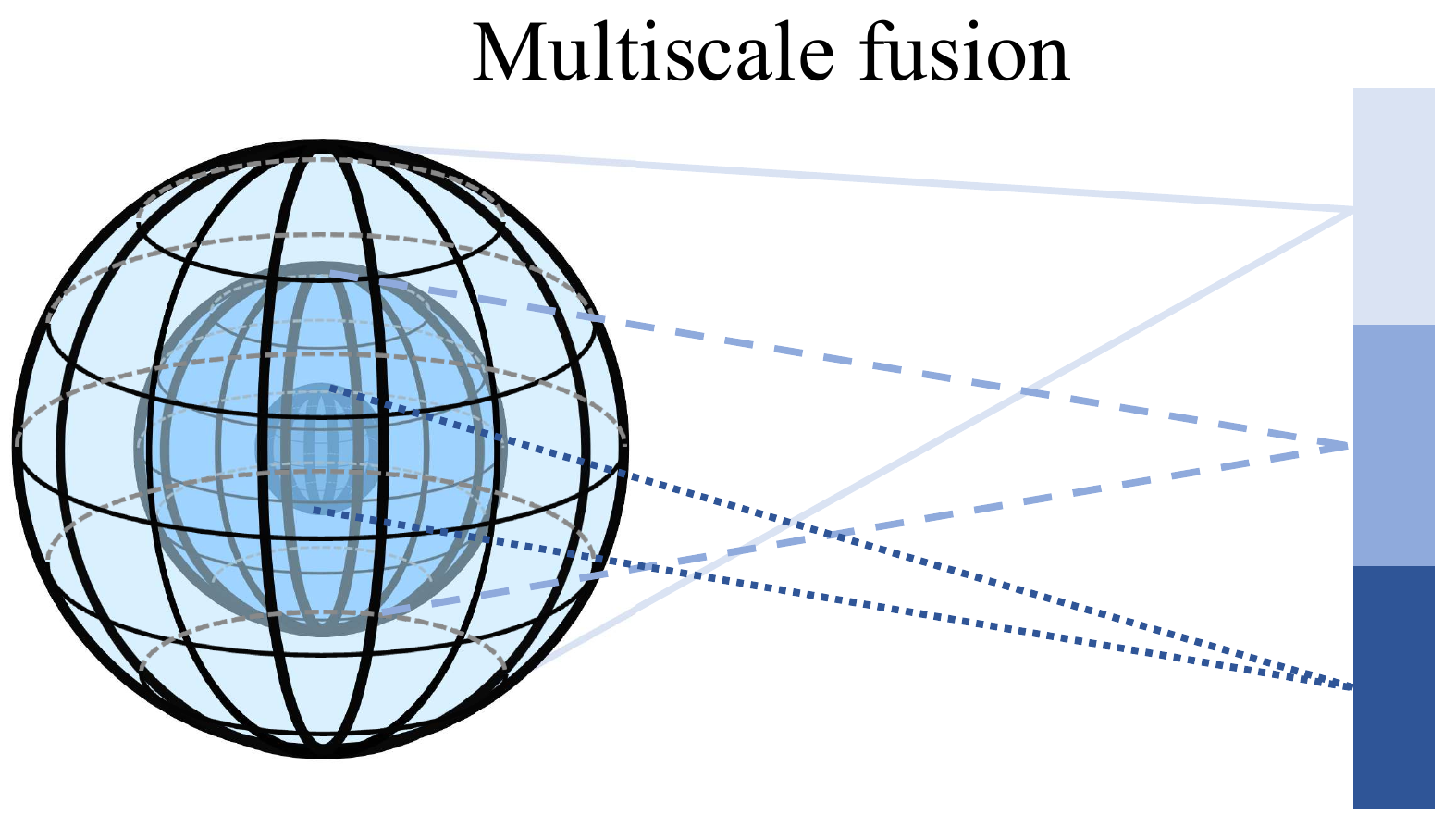}}  
   \caption{Illustration of multiscale feature fusion.}
   \label{MSG}

\end{figure}

\ptitle{Multiscale Feature Fusion.} 
In different datasets, the density of the point cloud is different. 
To resist the impact on density and improve the 
generalization ability of our SphereNet on cross-source datasets, 
we follow PointNet++~\cite{qi2017pointnet++} and 
perform multiscale  feature fusion on our spherical voxels.
As shown in Figure \ref{MSG}, 
we use three radii to obtain three spherical voxels 
at different scales. 
Then, the three are concatenated on 
the feature channel to complete 
the multiscale  fusion, which can be described as a mapping
$\mathcal M(\cdot): \mathbb{R}^{N \times M \times K} \rightarrow \mathbb{R}^{3 \times \frac{N}{3} \times M \times K}$.

\subsection{Spherical Feature Extractor}
For spherical feature extractor, we don't follow
spherical CNN~\cite{cohen2018spherical} or SpinNet~\cite{ao2021spinnet}. Instead, we use a simple
and efficient 3D sphere flattening to form 2D map, and then
carried out 2D convolution with a sphere, which is very suitable for deep extraction of our initial features. 
Compared with recent methods~\cite{huang2021predator,yu2021cofinet,qin2022geometric} that take KPConv~\cite{thomas2019kpconv} as a backbone, 
our network design provides a new idea for unseen-general and noise-robust 
descriptor extraction.

\begin{figure}[htbp]
\centering{\includegraphics[scale=0.23]{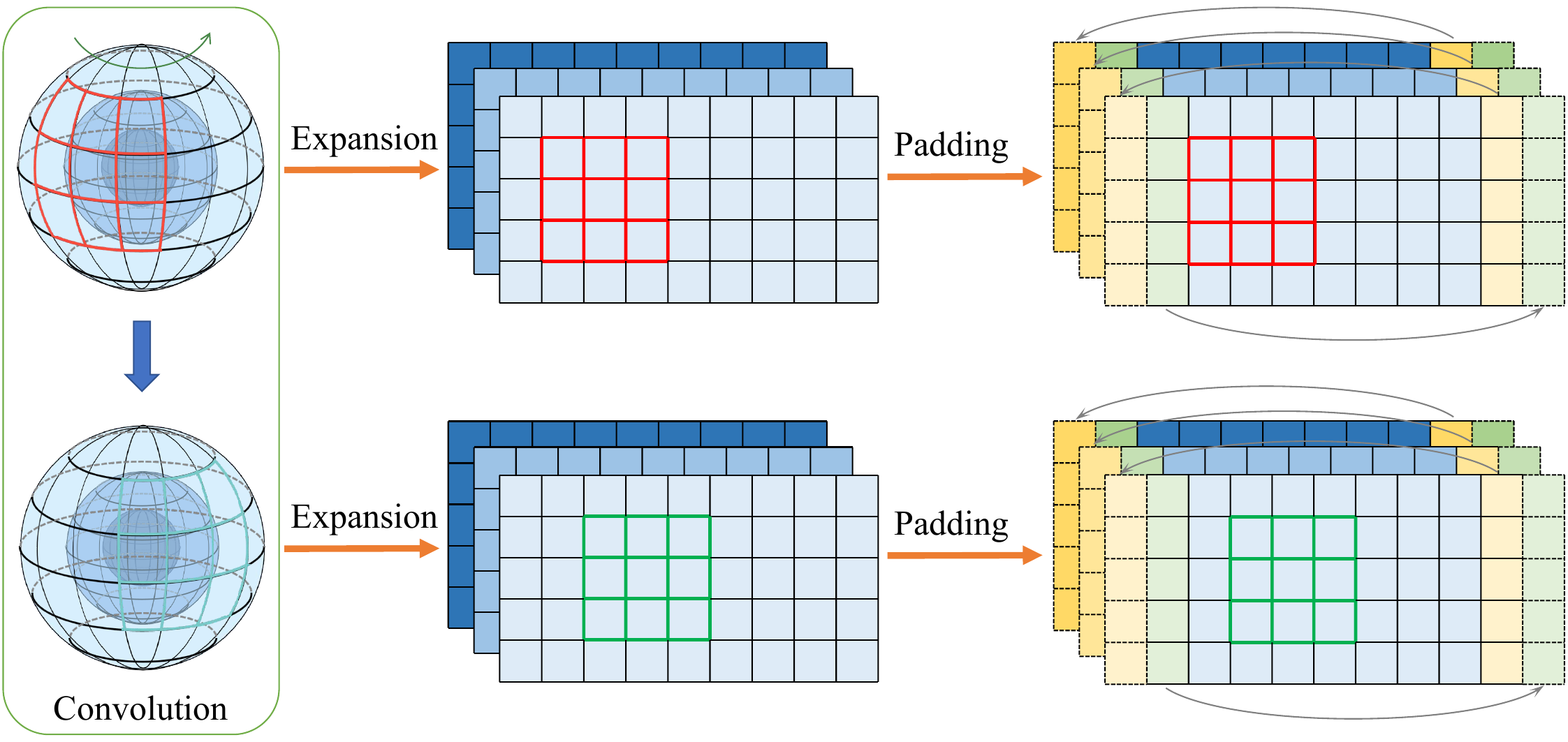}}  
\caption{Spherical convolution and boundary padding.}
\label{Sconvalution}
\end{figure}

\ptitle{Spherical Convolution.} \label{Spherical Convolution}
Before our work, 
many scholars have done research 
\cite{cohen2018spherical,esteves2018learning,jiang2019spherical,lei2019octree} on spherical convolution. 
Huan \etal \cite{lei2019octree} propose an Octree-guided CNN by combining CNN with Octree. 
And Taco \etal \cite{cohen2018spherical} 
pioneered the feasibility of spherical convolutions
and applied FFT to spherical convolutions. 
Different from the two methods, our SphereNet creatively 
proposes a new 3D spherical convolution and spherical 
integrity padding method, 
whose working principle can be compared with the 
projection expansion of the earth map, 
flattening the voxel sphere and obtaining a 
3D tensor similar to a cuboid as shown in Figure \ref{Sconvalution}. 
Therefore, convolution is carried out on the flattened tensor
through a 3D convolution kernel. 

\ptitle{Spherical Integrity Padding.}
Because the voxel sphere is cut lengthwise at azimuth angle $\varphi =0^{\circ}$ 
to flatten the sphere, the left and right boundaries in the 3D tensor 
are furthest apart, while the boundaries 
are adjacent to each other in the 
spherical voxels, which is bound 
to lead to the loss of geometric information. 
Therefore, we propose a spherical integrity padding 
to ensure that complete geometric features 
can be extracted from the boundaries.
As shown in Figure~\ref{Sconvalution}, 
the left and right boundaries after spherical flattening 
are padded with each other every time 3D convolution.

\ptitle{Spherical CNN.} \label{SCNN}
In the spheroid generator, we have already discretized a large number 
of points into the sphere with many small voxels. 
The ordered voxel sphere is very suitable for the implementation 
of the spherical convolutional neural network (Spherical CNN). 
By Spherical CNN, the mapping $\mathcal{F}(\cdot)$ between the input and output feature maps 
$\boldsymbol{h}$ of each layer is shown in the following equation:
\begin{equation}
\mathcal{F} (\boldsymbol{h}_{nmk}^{d})=\sum_{d=1}^D \sum_{r=1}^{a} \sum_{\theta =1}^{b} \sum_{\varphi =1}^{c} \boldsymbol{\omega}_{r \theta \varphi }^{d^{\prime}d} \boldsymbol{h}_{(n+r)(m+\theta )(k+\varphi )}^{d}
\label{eq:SCNN}
\end{equation}
where $a$, $b$, $c$, $D$ are the size and number of the convolution 
filter $\Omega \in \mathbb{R}^{a \times b \times c}$
, and $\boldsymbol{\omega}_{r \theta \varphi }^{d^{\prime}d}$ is a learnable parameter in the convolution kernel where $d$ and $d^{\prime}$ are 
the number of input and output channels, respectively. 
The rotation invariance of the end-to-end model is proved in \ref{lemma4}.

\begin{theorem} \label{lemma3}
\emph{
   Given any XOY-plane rotation transformation $R_{xy|z_n} \in SO(2)$ 
   fixed on the direction $\mathbf{z}_n$, then the Spherical CNN is 
   an equivariant mapping $\mathcal{F}(\cdot)$ of the transformation $R_{xy|z_n}$.
}
\end{theorem}
\begin{IEEEproof}[Proof of Theorem \ref{lemma3}]
   \emph{
   According to Eq. 11 (in our paper), the spherical CNN can regarded as a mapping $\mathcal{F}(\cdot)$
   that transforms from one feature map $\boldsymbol{h}_{nmk}^d$ to another $\boldsymbol{h}_{nmk}^{d^{\prime}}$.
   When the transformation $R_{xy|z_n}$ 
   is performed on $\boldsymbol{h}_{nmk}^d$, since the transformation is only in the XOY-plane, 
   the feature $\boldsymbol{h}_{nmk}^d$ 
   will be transformed into $\boldsymbol{h}_{nm(k+\zeta)}$.
   Therefore, the equivariant mapping of the transformation $R_{xy|z_n}$ can be proved as follows:
   }
   \begin{equation}
   \mathcal{F} (R_{xy|z_n}\circ  \boldsymbol{h}_{nmk}^d)=\mathcal{F} (\boldsymbol{h}_{nm(k+\zeta)})
   =R_{xy|z_n}\circ \mathcal{F} (\boldsymbol{h}_{nmk}^d)
   \end{equation}
   
\end{IEEEproof}

\begin{theorem}\label{lemma4}
   \emph{
   Given any rotation transformation $R_{xyz}\in SO(3)$, 
   the end-to-end framework of feature extraction is an invariant mapping.
   }
   \end{theorem}
   
   \begin{IEEEproof}[Proof of Theorem \ref{lemma4}]
   \emph{
   Firstly, it is known that maximum pooling has invariance. 
   So we can get the following property:}
   \begin{equation}
      \mathcal{O}(\boldsymbol{h}_{nmk}^d) =\mathcal{O}(R_{xy|z_n} \circ  \boldsymbol{h}_{nmk}^d)
   \end{equation}
   \emph{
   Then, according to the properties we proved in Lemma \ref{lemma1}, \ref{lemma2}, \ref{lemma3} 
   we can derive the end-to-end framework of feature extraction step by step:}
   \begin{equation}
      \begin{aligned} 
      &(\mathcal{O} \star \mathcal{F} \star \mathcal{H} \star \mathcal{L} )(R_{xyz}\circ \mathbf{P}^{i}) \\
      &=\mathcal{O}(\mathcal{F}(\mathcal{H}(\mathcal{L}(R_{xyz}\circ \mathbf{P}^{i})))) \\
      &=\mathcal{O}(\mathcal{F}(\mathcal{H}(R_{xy|z_n} \circ\mathcal{L}(\mathbf{P}^{i})))) \\
      &=\mathcal{O}(\mathcal{F}(R_{xy|z_n} \circ\mathcal{H}(\mathcal{L}(\mathbf{P}^{i})))) \\
      &=\mathcal{O}(R_{xy|z_n} \circ\mathcal{F}(\mathcal{H}(\mathcal{L}(\mathbf{P}^{i})))) \\
      &=\mathcal{O}(\mathcal{F}(\mathcal{H}(\mathcal{L}(\mathbf{P}^{i})))) 
      \end{aligned} 
   \end{equation}
   \emph{
   where mapping $\mathcal{O} (\cdot)$ is the maximum pooling, 
   $(\mathcal{O} \star \mathcal{F} \star \mathcal{H} \star \mathcal{L} )(\cdot)$ is the mapping of the whole framework 
   and operation $\star$ is the composite product between two transformations.}
   \end{IEEEproof}



\subsection{Detailed Network Architecture}
\begin{figure*}[!t]
   \centering{\includegraphics[scale=0.9]{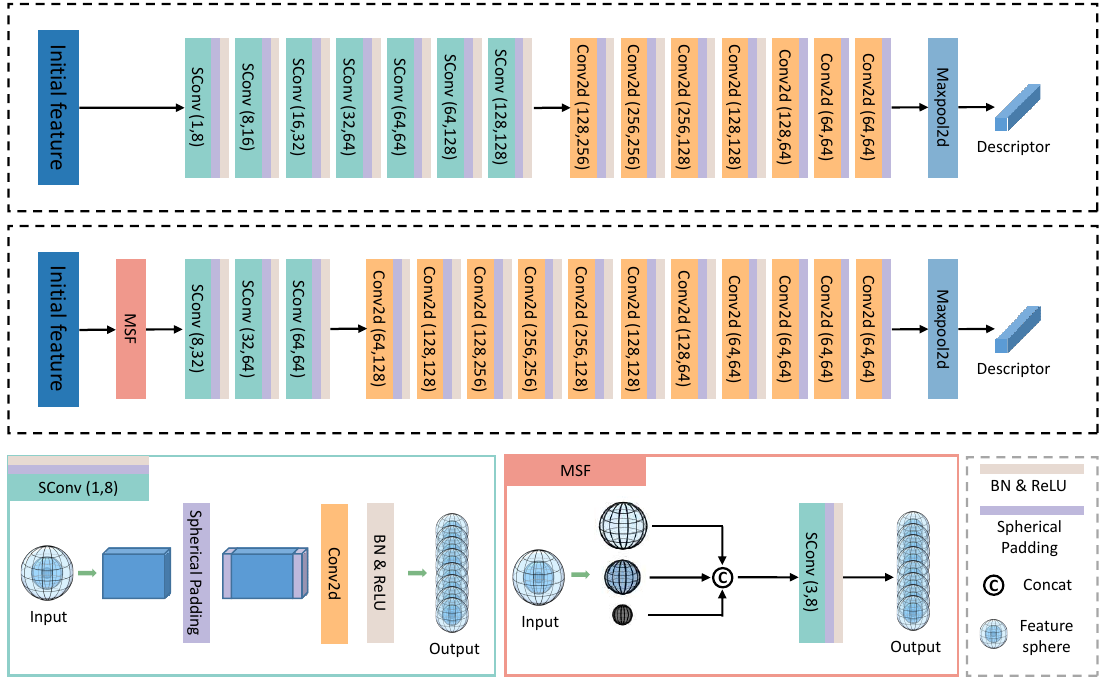}}  
   \caption{
      Detail network structure of our SphereNet.
   }
   \label{network}
\end{figure*}

The detail network structure of our SphereNet is shown in Figure \ref{network}. 
Our model is mainly based on Spherical convolution (SConv) proposed in this paper, 
and SConv combines with conv2d to form the network of our SphereNet. 
The spherical integrity padding is conducted before each convolution. 
As shown in Figure \ref{network}, we provide two different network structures. 
The first network (top) does not use the MSF module and is trained from the 3DMatch dataset for direct testing on the 3DMatch dataset. 
The second network (middle) uses MSF to improve the generalization ability of our model. 
After experimental comparative analysis, the second model has better generalization ability and can be used for generalization between different datasets.

\subsection{Loss Function}
Our method is patch-learning and we choose the \textit{hardest-in-batch} 
approach\cite{mishchuk2017working} for the
design of the loss function. 
During training, a batch $\mathcal{B}=(\mathcal{P},\mathcal{Q})$ is given by 
the combination of Source $\mathcal{P}$ and Target $\mathcal{Q}$, where the patch
$ \mathbf{P}_i \in \mathcal{P}$, patch $ \mathbf{Q}_i \in \mathcal{Q}$, and $(\mathbf{P}_i,\mathbf{Q}_i) \in \mathcal{G}$,
$(i=1,2,\cdots n)$. The \textit{hardest-in-batch} 
contrastive margin loss $\mathcal{L}_{cm}$ is defined as:
\begin{equation}
\begin{aligned}
   \mathcal{L}_{cm}&=\frac{1}{|\mathcal{P}|} \sum_{\mathbf{P}_i \in \mathcal{P}}\left(\max \left[\left(\max d^{\mathbf{h}}_{i}-\Delta_p\right), 0\right]\right. \\
   &\left.+ \max \left[\left(\Delta_n-\min d^{\mathbf{h}}_{ij}\right), 0\right] \right)
\end{aligned}
\end{equation}
where  $d^{\mathbf{h}}_{i}=\|\mathbf{h}_{\mathbf{P}_i}-\mathbf{h}_{\mathbf{Q}_i}\|_2 $ 
is the feature distance between $\mathbf{P}_i$ and $\mathbf{Q}_i$ and 
$d^{\mathbf{h}}_{ij}=\|\mathbf{h}_{\mathbf{P}_i}-\mathbf{h}_{\mathbf{Q}_j}\|_2 $ 
is the feature distance between $\mathbf{P}_i$ and $\mathbf{Q}_j$.
$\Delta_p$ and $\Delta_n$ is the positive and negative margin of the feature distance, respectively.

\begin{figure*}[!t]
   \centering{\includegraphics[scale=0.63]{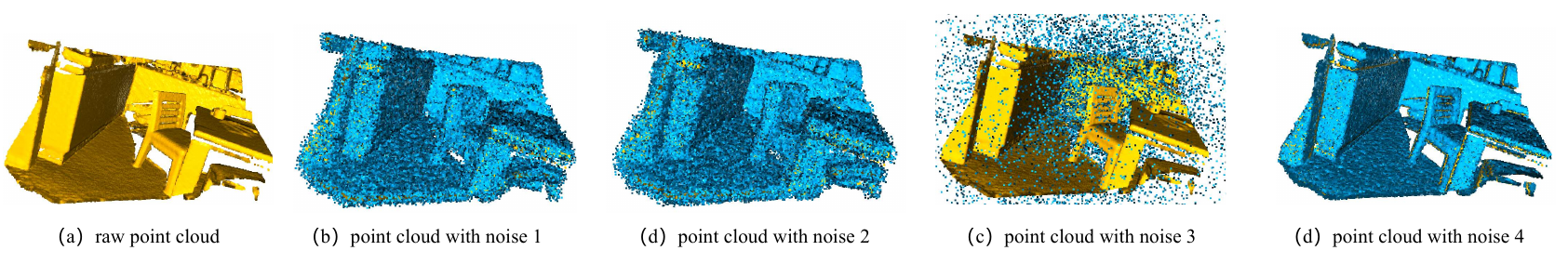}}  
   \caption{Examples in the 3DMatch-noise benchmark with different noises.
   The yellow point clouds are the original data of the
   3DMatch dataset, and the blue point clouds are the noise point clouds in the 3DMatch-noise benchmark.
   }
   \label{noisedata}
\end{figure*}

\section{Experiments}\label{Experiments}
This paper evaluates on two distinct datasets for indoor and outdoor scenes. 
For indoor scenes, we used the RGB-D reconstructed dataset 3DMatch\cite{zeng20173dmatch} 
to verify that our method achieved the best performance in indoor scenes. 
It is also evaluated on the KITTI \cite{geiger2012we} 
and ETH \cite{pomerleau2012challenging} datasets. 
Our approach is also competitive in outdoor scenes. 
It is worth noting that we also evaluate our SphereNet 
on the 3DMatch-noise benchmark proposed in this paper. 



\subsection{Experimental Setup}
First, we downsample the point cloud and generate patches as the input of 
our method. Then, we can use our spheroid generator and spherical feature extractor to extract the 3D feature 
descriptors. Finally, the extracted descriptors are used for registration by using 50k RANSAC. 
We implement our SphereNet in the PyTorch framework 
and choose Adam as our optimizer. We set the initial value of the learning rate to 0.001 
and decrease it by 50\% with each 5 epochs. 
The batch size and training epoch are set to 64 and 30, respectively. 
Finally, one of the training models with the best performance in validation
is selected as our pretraining model. 
All experiments are conducted on the Arch Linux system with an NVIDIA RTX 2080Ti GPU.

\ptitle{Hyperparameter Settings.}
Here are the detailed hyperparameter settings in this section. 
For different datasets and different experiments, the hyperparameters 
set in the experiment are given in Table \ref{tab:hyperparameter}. 
It is worth noting that in 
all experiments, we set the number of spherical voxels as $15\times 20\times 40 $. However, in order to overcome the scale 
inconsistency between different datasets, we adjusted the radius threshold $R$ 
of constructing LRFs
through the ablation experiment. For 3DMatch dataset, 
$R$ is to 0.3m because the indoor point clouds are dense. For the 
ETH and KITTI datasets, respectively, because the outdoor point clouds are sparse, 
$R$ are set to 1.2m and 3.0m. In addition, in order to 
improve the generalization ability of cross-dataset registration, 
multiscale  feature fusion (MSF) is used in some generalization experiments.

\begin{table}[htbp]
   \caption{Hyperparameter settings of our SphereNet.}
   \setlength{\tabcolsep}{7pt}
   \scriptsize
   \centering
   \begin{tabular}{lcccccc}
   \toprule
   Dataset & $N$ & $M$  & $K$  & $R$    & MSF \\
   \midrule
   3DMatch & 15 & 20 & 40 & 0.3m & NO   \\
   KITTI    & 15 & 20 & 40 & 2.0m  & NO   \\
   3DMatchToETH  & 15 & 20 & 40 & 1.2m & YES   \\
   3DMatchToKITTI  & 15 & 20 & 40 & 3.0m  & YES   \\
   \bottomrule
   \end{tabular}
   \label{tab:hyperparameter}
   \end{table}

\subsection{Evaluation on 3DMatch \& 3DLoMatch}\label{Evaluation on 3DMatch Dataset}
\ptitle{Dataset and Metrics.}
Following previous works 
\cite{bai2020d3feat,ao2021spinnet}, 
we used 46 scenes for training, 8 scenes for validation, 
and 8 scenes for testing. On 3DMatch, 
we evaluate on three benchmarks: (1) the 3DMatch benchmark is tested on 
scenes with more than 30\% overlap on the 3DMatch dataset; 
(2) the 3DLoMatch benchmark is tested on scenes with $10\%\thicksim30\%$ overlap;
(3) the 3DMatch-noise benchmark is tested on the 3DMatch dataset with strong noise.

Following \cite{bai2020d3feat,yu2021cofinet}, we evaluated the 
performance by three evaluation metrics: \emph{Registration Recall} (RR), 
\emph{Inlier Ratio} (IR), and \emph{Feature Matching Recall} (FMR).

\emph{Feature Matching Recall} is the fraction of two point clouds whose 
\emph{Inlier Ratio} (IR) is above the threshold $\tau_2=5\%$.
So we should calculate IR before FMR.
First, we obtain the point pair correspondences $\left(\mathbf{p}_i, \mathbf{q}_j\right) \in \boldsymbol{\Omega}_h$
between two point clouds $\mathbf{P}$ and $\mathbf{Q}$ by feature matching. 
Then, $\mathbf{P}$ and $\mathbf{Q}$ are aligned according to the ground-truth 
transformation $\mathbf{T_h}\{\mathbf{R_h},\mathbf{t_h}\}$. 
Finally, the Euclidean distance between the aligned point pair $\left(\mathbf{p}_i, \mathbf{q}_j\right)$ is calculated. 
If the distance is less than the threshold $\tau_1=0.1$m, the point pair $\left(\mathbf{p}_i, \mathbf{q}_j\right)$ 
is considered as inlier point pair.
IR is the ratio of the number of inlier point pairs to the number of correspondences:
\begin{equation}
   \mathrm{IR}=\frac{1}{\left|\boldsymbol{\Omega}_h\right|} \sum_{\left(\mathbf{p}_i, \mathbf{q}_j\right) \in \boldsymbol{\Omega}_h} \mathbbm{1}\left(\left\|\mathbf{T_h}(\mathbf{p}_i)-\mathbf{q}_j\right\|_2<\tau_1\right)
\end{equation}
where $\left|\boldsymbol{\Omega}_h\right|$ is the number of correspondences and 
$\mathbbm{1}$ is an indicator function. $\mathbf{T}(\mathbf{p}_i)$ is the point through the ground-truth 
transformation $\mathbf{T_h}$ from $\mathbf{p}_i$. $\|\cdot\|_2$ denotes the Euclidean distance.

When the IR between two point clouds is above the threshold $\tau_2=5\%$, 
we believe that the feature matching of two point clouds is successful. 
Finally, FMR is the percentage of point clouds that IR exceeds the threshold $\tau_2=5\%$:
\begin{equation}
   \mathrm{FMR}=\frac{1}{H} \sum_{h=1}^H \mathbbm{1}\left(\mathrm{IR_h}>\tau_2\right)
\end{equation}
where $H$ is the total number of point cloud pairs.

\emph{Registration Recall} is the fraction of correctly registered point cloud pairs. 
First, we should calculate the RMSE which determines whether the registration of two point clouds $\mathbf{P}$ and $\mathbf{Q}$ is successful.
\begin{equation}
\operatorname{RMSE}=\sqrt{\frac{1}{\left|\boldsymbol{\Omega}_h\right|} \sum_{\left(\mathbf{p}_i, \mathbf{q}_j\right)  \in \boldsymbol{\Omega}_h}\left\|\mathbf{T_h}(\mathbf{p}_i)-\mathbf{q}_j\right\|_2^2}
\end{equation}
If the RMSE is less than the threshold $\tau_3=0.2$m, 
the registration of point clouds $\mathbf{P}$ and $\mathbf{Q}$ is regarded as successful. 
Therefore, RR is calculated by the following equation:
\begin{equation}
\mathrm{RR}=\frac{1}{H} \sum_{h=1}^H \mathbbm{1} \left( \mathrm{RMSE}_h<\tau_3 \right)
\end{equation}

\begin{table}[htbp]
   \caption{
      Evaluation results on 3DMatch and 3DLoMatch.
      }
   \setlength{\tabcolsep}{2.8pt}
   \scriptsize
   \centering
   \begin{tabular}{l|ccccc|ccccc}
   \toprule
      & \multicolumn{5}{c|}{3DMatch} & \multicolumn{5}{c}{3DLoMatch} \\
   \# Samples & 5000 & 2500 & 1000 & 500 & 250 & 5000 & 2500 & 1000 & 500 & 250 \\
   \midrule
   \multicolumn{11}{c}{\emph{Feature Matching Recall} (\%) $\uparrow$} \\
   \midrule
   PerfectMatch~\cite{gojcic2019perfect} & 95.0 & 94.3 & 92.9 & 90.1 & 82.9 & 63.6 & 61.7 & 53.6 & 45.2 & 34.2 \\
   FCGF~\cite{choy2019fully} & 97.4 & 97.3 & 97.0 & 96.7 & 96.6 & 76.6 & 75.4 & 74.2 & 71.7 & 67.3 \\
   D3Feat~\cite{bai2020d3feat} & 95.6 & 95.4 & 94.5 & 94.1 & 93.1 & 67.3 & 66.7 & 67.0 & 66.7 & 66.5 \\
   SpinNet~\cite{ao2021spinnet} & 97.6 & 97.2 & 96.8 & 95.5 & 94.3 & 75.3 & 74.9 & 72.5 & 70.0 & 63.6 \\
   Predator~\cite{huang2021predator} & 96.6 & 96.6 & 96.5 & 96.3 & 96.5 & 78.6 & 77.4 & 76.3 & {75.7} & 75.3 \\
   YOHO~\cite{wang2021you} & \underline{98.2} & 97.6 & 97.5 & 97.7 & 96.0 & 79.4 & 78.1 & 76.3 & 73.8 & 69.1 \\
   CoFiNet \cite{yu2021cofinet} & 98.1 & \textbf{98.3} & {98.1} & \underline{98.2} & \textbf{98.3} & {83.1} & {83.5} & {83.3} & \underline{83.1} & \underline{82.6} \\ 
   GeoTransformer \cite{qin2022geometric} & 97.9 & {97.9} & {97.9} & {97.9} & {97.6} & \underline{88.3} & \textbf{88.6} & \textbf{88.8} & \textbf{88.6} & \textbf{88.3} \\ 
   SphereNet$^1$ (\emph{ours}) & \underline{98.2} & \textbf{98.3} & \underline{98.2} & 97.3 & 95.8 &79.1 &78.5&77.3&74.1&68.4 \\ 
   SphereNet$^2$ (\emph{ours}) & \textbf{98.5} & \textbf{98.3} & \textbf{98.3} & \textbf{98.3} & \underline{98.0} & \textbf{88.8} & \underline{87.0} & \underline{86.2} & \underline{83.1} & {82.0} \\ 
   \midrule
   \multicolumn{11}{c}{\emph{Registration Recall} (\%) $\uparrow$} \\
   \midrule
   PerfectMatch~\cite{gojcic2019perfect} & 78.4 & 76.2 & 71.4 & 67.6 & 50.8 & 33.0 & 29.0 & 23.3 & 17.0 & 11.0 \\
   FCGF~\cite{choy2019fully} & 85.1 & 84.7 & 83.3 & 81.6 & 71.4 & 40.1 & 41.7 & 38.2 & 35.4 & 26.8  \\
   D3Feat~\cite{bai2020d3feat} & 81.6 & 84.5 & 83.4 & 82.4 & 77.9 & 37.2 & 42.7 & 46.9 & 43.8 & 39.1 \\
   SpinNet~\cite{ao2021spinnet} & 88.6 & 86.6 & 85.5 & 83.5 & 70.2 & 59.8 & 54.9 & 48.3 & 39.8 & 26.8 \\
   Predator~\cite{huang2021predator} & 89.0 & 89.9 & {90.6} & 88.5 & 86.6 & 59.8 & 61.2 & 62.4 & 60.8 & 58.1 \\
   YOHO~\cite{wang2021you} & {90.8} & {90.3} & 89.1 & {88.6} & 84.5 & 65.2 & 65.5 & 63.2 & 56.5 & 48.0 \\
   CoFiNet \cite{yu2021cofinet} & 89.1 & 88.9 & {88.4} & {87.4} & {87.0} & {67.5} & {66.2} & {64.2} & {63.1} & {61.0} \\ 
   GeoTransformer \cite{qin2022geometric} & \underline{92.0} & \underline{91.8} & \underline{91.8} & \underline{91.4} & \textbf{91.2} & \underline{75.0} & \textbf{74.8} & \textbf{74.2} & \textbf{74.1} & \textbf{73.5} \\
   SphereNet$^1$ (\emph{ours}) & 91.2 & 90.3 & 88.8 & 86.2 & 77.7  &60.0 & 59.6&53.9&45.5&32.4 \\
   SphereNet$^2$ (\emph{ours}) & \textbf{93.4} & \textbf{93.5} & \textbf{93.0} & \textbf{92.4} & \underline{89.8} & \textbf{75.6} & \underline{71.2} & \underline{70.0} & \underline{70.6} & \underline{65.9} \\
   \bottomrule
   \end{tabular}
   \label{table:results-3dmatch}
   \end{table}

\ptitle{Correspondence Results.}
Following \cite{yu2021cofinet,ao2021spinnet}, 
we first compare the correspondence results of our method 
with the state-of-the-art methods \cite{gojcic2019perfect,choy2019fully,bai2020d3feat,ao2021spinnet,huang2021predator,wang2021you,
ao2022YOTO,yu2021cofinet} in Table \ref{table:results-3dmatch}.
To verify the robustness to samplings, 
we set up experiments with sampling numbers 5000, 2500, 1000, 500 and 250. 
For keypoint sampling, we provide two versions of SphereNet. 
The first is our SphereNet$^1$, which uses random downsampling 
to obtain keypoints; the second is SphereNet$^2$, 
where we add the keypoint detection module of Predator \cite{huang2021predator} into our framework to obtain keypoints.
For FMR, our SphereNet$^2$ achieves the best FMR scores of 98.5\% and 88.8\% on 3DMatch and 3DLoMatch, respectively.
Additionally, our SphereNet$^1$ without keypoint detection or coarse-to-fine 
matching \cite{yu2021cofinet,qin2022geometric} also achieves the highest FMR than the methods \cite{huang2021predator, ao2022YOTO, yu2021cofinet}.

\ptitle{Registration Results.}
50k RANSAC is used to calculate the transformation through feature matching. 
Notably, whether on the 3DMatch or 3DLoMatch dataset, our SphereNet$^2$ has the best performance for any number of keypoints.
It attains a new state-of-the-art result of 93.4\%, 
a 4.8 percentage points (pp) increase compared with baseline SpinNet \cite{ao2021spinnet} on 3DMatch. 
For 3DLoMatch,
our SphereNet$^2$ also achieves the best RR score of 75.6\%.
Moreover, with the decrease in the sampling numbers, 
the decrease in RR of our method is small, which means that our method has good stability under different samplings.
As the qualitative results on 3DMatch shown in Figure~\ref{Qualitative1}, our SphereNet achieves
successful registration in some examples where other 
methods~\cite{huang2021predator,yu2021cofinet} fail.

We also conduct detailed experiments 
for each scene under the 3DMatch and 3DMatch-noise datasets (introduced in Section \ref{3DMatch-noise}). 
Following \cite{ao2021spinnet}, 
we calculated RR for each scene
to evaluate the performance, and the results are 
shown in Table \ref{table:scene-wise}. 
As can be seen from the Table \ref{table:scene-wise}, 
compared with the baselines, 
the performance of our method in each scene has 
been greatly improved. 
In particular, due to the characteristics of our method, 
the performance is significantly improved
in all scenes on the 3DMatch-noise dataset with strong noise.

\begin{table*}[!t]
   \caption{
      Scene-wise registration results on the 3DMatch and 3DMatch-noise datasets.
      }
   \scriptsize
   \setlength{\tabcolsep}{2pt}
   \centering
   \begin{tabular}{l|ccccccccc|ccccccccc}
   \toprule
   \multirow{2}{*}{Model} & \multicolumn{9}{c|}{3DMatch} & \multicolumn{9}{c}{3DMatch-noise} \\
    & Kitchen & Home\_1 & Home\_2 & Hotel\_1 & Hotel\_2 & Hotel\_3 & Study & Lab & Mean & Kitchen & Home\_1 & Home\_2 & Hotel\_1 & Hotel\_2 & Hotel\_3 & Study & Lab & Mean \\
   \midrule
   \multicolumn{19}{c}{\emph{Registration Recall} (\%) $\uparrow$} \\
   \midrule
   3DSN~\cite{gojcic2019perfect} & 90.6 & 90.6 & 65.4 & 89.6 & 82.1 & 80.8 & 68.4 & 60.0 & 78.4 & - & - & - & - & - & - & - & - & -  \\
   FCGF~\cite{choy2019fully} & {98.0} & 94.3 & 68.6 & 96.7 & {91.0} & \underline{84.6} & 76.1 & 71.1 & 85.1 & - & - & - & - & - & - & - & - & -  \\
   D3Feat~\cite{bai2020d3feat} & 96.0 & 86.8 & 67.3 & 90.7 & 88.5 & 80.8 & 78.2 & 64.4 & 81.6 & - & - & - & - & - & - & - & - & -  \\
   Predator~\cite{huang2021predator} & 97.6 & \underline{97.2} & {74.8} & \underline{98.9} & \textbf{96.2} & \textbf{88.5} & 85.9 & 73.3 & 89.0 &42.3  &41.5  &28.9  &48.4  &39.7  &61.5  &45.3  &46.7  & 44.3  \\
   CoFiNet~\cite{yu2021cofinet} & 96.4 & \textbf{99.1} & 73.6 & {95.6} & {91.0} & \underline{84.6} & \textbf{89.7} & \underline{84.4} & {89.3} &\underline{57.2}  &58.5  &39.0  &\underline{65.4}  &53.8  &50.0  &51.3  &46.7  &52.7  \\
   GeoTransformer~\cite{qin2022geometric} & \textbf{98.9} & \underline{97.2} & \textbf{81.1} & \underline{98.9} & 89.7 & \textbf{88.5} & \underline{88.9} & \textbf{88.9} & \underline{91.5} &56.3  &57.6  &40.3  &65.2  &\underline{55.2}  &52.6  &44.9  &41.5  &51.7  \\
   SphereNet$^1$ (\emph{ours}) & \underline{98.2} & {95.3} & {74.2} & \textbf{99.5} & \underline{93.6} & {92.3} & {82.9} & 75.6 & {91.2} & {55.5} & \underline{59.4} & \underline{43.4} & {56.6} & {43.6} & \underline{57.7} & \underline{57.3} & \underline{48.9} & \underline{53.8} \\
   SphereNet$^2$ (\emph{ours}) & \textbf{98.9} & {96.2} & \underline{80.5} & \underline{98.9} & \underline{93.6} & \textbf{88.5} & {87.6} & \textbf{88.9} & \textbf{93.4} & \textbf{78.6} & \textbf{82.1} & \textbf{55.4} & \textbf{81.9} & \textbf{66.7} & \textbf{73.1} & \textbf{78.2} & \textbf{60.0} & \textbf{74.9} \\
   \bottomrule
   \end{tabular}
   \label{table:scene-wise}
   \end{table*}

\begin{figure}[htbp]
   \centering{\includegraphics[scale=0.44]{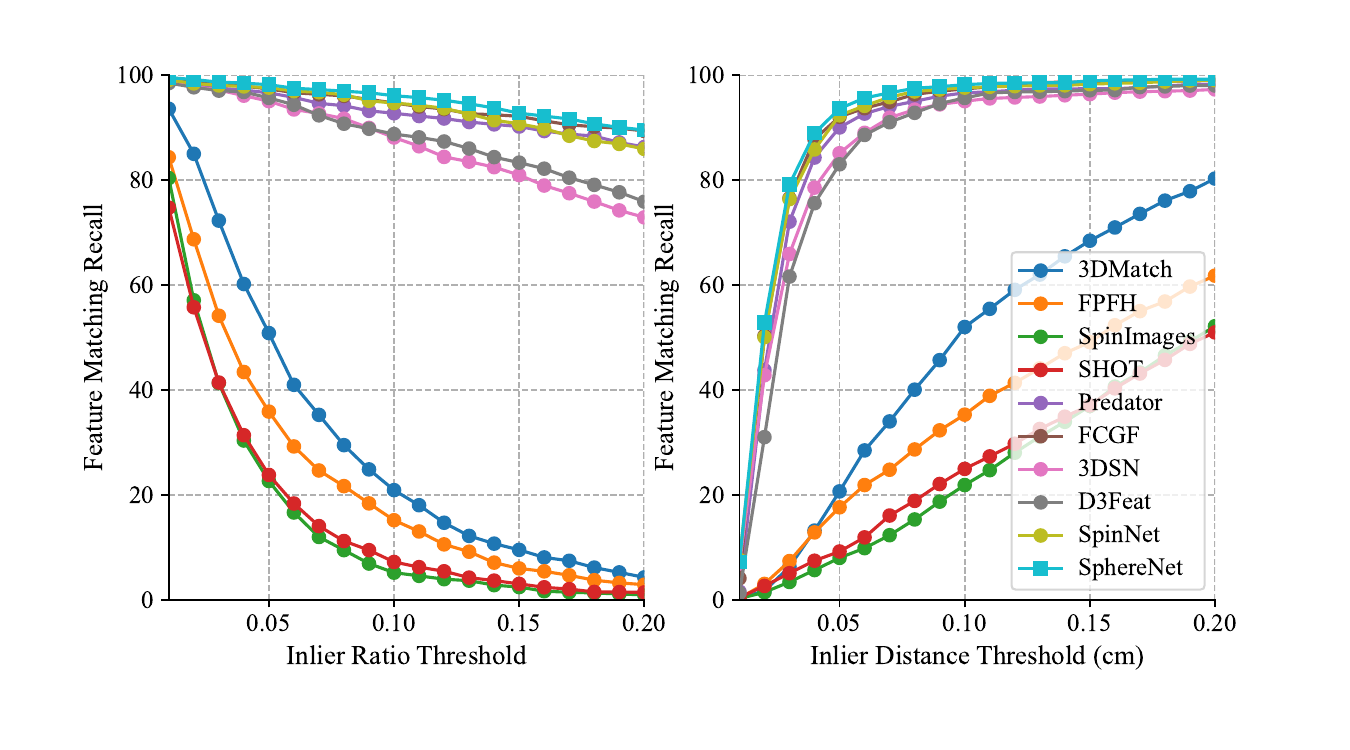}}
	\caption{Feature Matching Recall on the 3DMatch dataset under different inlier distance threshold $\tau_1$ and inlier ratio threshold $\tau_2$.}
	\label{fig:3dmatch_FMR}
\end{figure}

\ptitle{Repeatability for Different Thresholds.}
We also set different thresholds to verify the robustness to 
the thresholds $\tau_1$ and $\tau_2$. As shown in Figure \ref{fig:3dmatch_FMR}, 
compared with other methods~\cite{huang2021predator,ao2021spinnet,FPFH,salti2014shot,zeng20173dmatch,bai2020d3feat}, 
our SphereNet achieves the best results at all different thresholds. 
Under the stricter condition $\tau_1=0.2$, 
the FMR obtained by our SphereNet is $89.5\%$, 
while other methods~\cite{zeng20173dmatch,bai2020d3feat} 
have a significant decline. 
Therefore, our method has good stability and 
can achieve good results even in the face of strict conditions.

\ptitle{Rotation Invariance.}
Following \cite{bai2020d3feat,ao2021spinnet},
we build a rotated 3DMatch benchmark 
by applying arbitrary rotations in SO(3) 
group to all fragments of the 3DMatch dataset. 
On the rotated 3DMatch benchmark, 
the FMR scores are calculated by our method and 
other strong baselines~\cite{choy2019fully, bai2020d3feat,wang2021you,qin2022geometric,ao2021spinnet} 
to verify the rotation invariance of our SphereNet. 
As is shown in Table~\ref{tab:3dmatch}, our SphereNet achieves the highest average FMR 
score and the lowest standard deviation on the original and rotated
datasets. Notably, SphereNet achieves good rotation invariance 
without the use of data augmentation and keypoint detection.

\begin{table}[htb]
   \caption{Results for rotation invariance on the 3DMatch dataset, where Key.Det.:keypoint detection, Rot.Aug.:rotation augmentation.
    }
   \setlength{\tabcolsep}{4pt}
   \scriptsize
   \centering
			\begin{tabular}{l|cc|cc|c|c}
		      \toprule
				\multirow{2}{*}{} & \multicolumn{2}{c|}{\bf{Origin}}  & \multicolumn{2}{c|}{\bf{Rotated}} & {Key.} & {Rot.} \\
				& FMR (\%)      & STD         & FMR (\%)      & STD   & {Det.}   & {Aug.}    \\ \midrule
				PerfectMatch~\cite{gojcic2019perfect} & 94.7          & 2.7          & 94.9          & 2.5    & {No}  & {No}    \\
				FCGF~\cite{choy2019fully}         & 95.2          & 2.9          & 95.3          & 3.3    & {No}      & {Yes}    \\
				D3Feat~\cite{bai2020d3feat} & 95.3          & 2.7          & 95.2          & 3.2   & {Yes}    & {Yes}     \\
            SpinNet~\cite{ao2021spinnet} & 97.6          & \underline{1.9}          & 97.5          & \textbf{1.9}   & {No}   & {No}     \\
            D3Feat~\cite{huang2021predator} & 96.4          & 2.4          & 94.1          & 3.9   & {Yes}    & {Yes}     \\
				YOHO~\cite{wang2021you}          & \underline{98.2}          &  2.4       &  \underline{98.1}         &2.8    & {No}   & {No}   \\
            YOTO~\cite{ao2022YOTO}          & 97.8          & \textbf{ 1.7}       &  97.8         &2.0    & {No}   & {No}   \\
            CoFiNet~\cite{yu2021cofinet}          & 98.1          &  2.6       &  97.4         &2.9    & {No}   & {No}   \\
            GeoTransformer~\cite{qin2022geometric} &97.9      & 2.0          & 97.7          & \underline{2.3}   & {No}     & {No}     \\
				\bf SphereNet (\emph{ours})        & \textbf{98.3} & \textbf{1.7} & \textbf{98.3} & \textbf{1.9} & \textbf{No} & {\textbf{No}} \\ 
		      \bottomrule
			\end{tabular}
	\label{tab:3dmatch}

\end{table}

\begin{table*}[htb]
   \caption{Results on the 3DMatch-noise benchmark. Noise 1, Noise 2, Noise 3 and Noise 4 are 
   four kinds of noise in our 3DMatch-noise benchmark.}
   \scriptsize
   \centering
   \begin{tabular}{l|ccc|ccc|ccc|ccc|ccc|c}
      \toprule
      \multirow{2}{*}{} & \multicolumn{3}{c|}{\bf{Origin}}& \multicolumn{3}{c|}{\bf{Noise 1}}  & \multicolumn{3}{c|}{\bf{Noise 2}} &\multicolumn{3}{c|}{\bf{Noise 3}} & \multicolumn{3}{c|}{\bf{Noise 4}} &\bf{Noise}   \\
      & FMR    &IR   & RR        & FMR    &IR   & RR        & FMR  &IR    & RR    & FMR   &IR    & RR  & FMR   &IR    & RR &\bf{Augmentation}   \\ 
      \midrule
      Predator~\cite{huang2021predator} &96.6&58.0&89.0 &28.3&3.8&44.3 &50.7&6.9&55.6 &84.6&29.6&{88.1} &76.4&81.3&28.9    &Yes\\
      CoFiNet~\cite{yu2021cofinet} &98.1&49.8&89.3 &65.3&10.5&52.7 &81.6&17.4&\underline{65.7} &96.0&38.2&85.3 &78.6&93.3&33.9 &Yes\\
      Lepard~\cite{li2022lepard} &98.0&57.6&80.0 &59.4&9.5&14.4 &77.7&15.7&26.4 &94.5&27.9&41.7 &75.2&90.3&25.9 &Yes\\
      GeoTransformer~\cite{qin2022geometric} &97.9&\textbf{71.9}&\underline{92.0} &59.8&8.9&51.7 &80.3&16.2&61.0 &95.3&42.3&86.2 &79.0&92.1&65.1 &Yes\\
      SphereNet$^1$  (\emph{ours})  &\underline{98.2}&52.6&91.2 &\underline{85.5}&\underline{32.2}&\underline{66.4} &\underline{87.5}&\underline{35.6}&{67.3}  &\underline{96.7}&\underline{43.2}&\underline{88.2} &\underline{94.1}&\underline{39.9}&\underline{82.1} &No\\
      SphereNet$^2$ (\emph{ours})  &\textbf{98.5}&\underline{67.3}&\textbf{93.4}    &\textbf{90.3}&\textbf{33.5}&\textbf{74.9}    &\textbf{93.7}&\textbf{39.7}&\textbf{81.6}  &\textbf{97.7}&\textbf{58.8}&\textbf{92.3} &\textbf{96.3}&\textbf{49.2}&\textbf{87.5} &No \\

      \bottomrule
   \end{tabular}

	\label{noise}
\end{table*}

\begin{figure*}[!t]
   \centering{\includegraphics[scale=0.85]{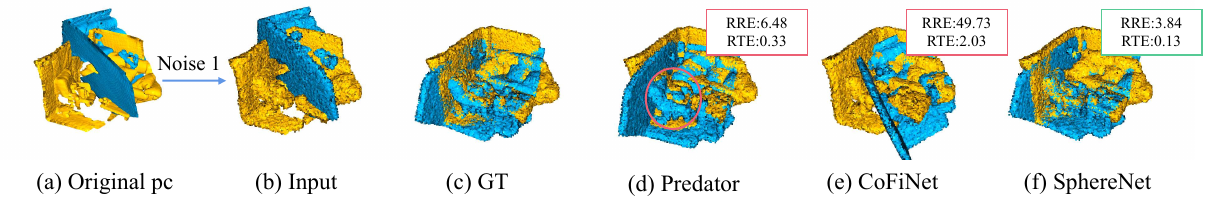}}  
   \caption{Qualitative results on the 3DMatch-noise benchmark. Our method achieves correct registration with the lowest RRE and RTE.
   }
   \label{qr}

\end{figure*}

\begin{table}[htbp]
   \caption{
      Registration results on KITTI odometry.
      }
   \scriptsize
   \centering
   \begin{tabular}{l|ccc}
   \toprule
   Method & \textbf{RTE(cm)} & \textbf{RRE($^{\circ}$)} & \textbf{RR(\%)} \\
   \midrule
   3DFeat-Net~\cite{yew20183dfeat} & 25.9 & \textbf{0.25} & 96.0 \\
   Predator~\cite{huang2021predator} & \textbf{6.8} & \underline{0.27} & \textbf{99.8} \\
   CoFiNet~\cite{yu2021cofinet} & 8.2 & 0.41 & \textbf{99.8} \\
   GeoTransformer~\cite{qin2022geometric} & \underline{7.4} & \underline{0.27} & \textbf{99.8} \\
   SphereNet (\emph{ours}) &\textbf{6.8}  &{0.31}  &\textbf{99.8}  \\
   \bottomrule
   \end{tabular}

   \label{table:kitti}
\end{table}

\subsection{Evaluation on KITTI}
\ptitle{Dataset and Metrics.}
Following  \cite{bai2020d3feat, choy2019fully, huang2021predator, ao2021spinnet},
we used sequences 0-5 for training, sequences 6-7 for validation, 
and sequences 8-10 for testing.
Since the ground truth transformation $\mathbf T\{\mathbf R,\mathbf t\}$ 
raised by the authors is not accurate, 
we follow \cite{bai2020d3feat, choy2019fully} 
to use the iterative closest point (ICP) \cite{besl1992method} algorithm to refine the ground truth 
transformation  $\mathbf T$.

Following previous works \cite{huang2021predator, ao2021spinnet}, 
we evaluate our SphereNet by three metrics:
\emph{Relative Rotation Error} (RRE), \emph{Relative Translation Error} (RTE), and
\emph{Success Rate} (SR). 

\emph{Relative Rotation Error}, the geodesic distance between estimated 
and ground-truth rotation matrices,
can be used to measure the error between our 
predicted rotation transformation and the ground-truth rotation transformation:
\begin{equation}
\mathrm{RRE}=\arccos \left(\frac{\operatorname{trace}\left(\hat{\mathbf{R}}_h^T \mathbf{R}_h\right)-1}{2}\right)
\end{equation}
where $\hat{\mathbf{R}}_h$ and $\mathbf{R}_h$ are our 
predicted rotation matrix and the ground-truth rotation matrix respectively.

\emph{Relative Translation Error}, the Euclidean distance between the estimated and ground-truth translation vectors, 
can be used to measure the error between our 
predicted translation transformation and the ground-truth translation transformation:
\begin{equation}
   \mathrm{RTE}=\left\|\hat{\mathbf{t}}_h-\mathbf{t}_h\right\|
\end{equation}
where $\hat{\mathbf{h}}_t$ and $\mathbf{h}_t$ are our 
predicted translation vector and the ground-truth translation vector respectively.

\emph{Success Rate} is the ratio of the number of registration that errors 
within certain thresholds, which determines whether the registration is successful on KITTI.
If the RRE is less than $2$m and the RTE is less than $5^{\circ}$,
the registration on this point cloud pair is considered a successful registration. 
\begin{equation}
\mathrm{SR}=\frac{1}{H} \sum_{h=1}^H \mathbbm{1}\left(\mathrm{RRE}_h<2 \mathrm{~m} \wedge  \mathrm{RTE}_h<5^{\circ}\right)
\end{equation}

\ptitle{Registration Results.}
We compare our SphereNet with strong baselines \cite{yew20183dfeat, huang2021predator,yu2021cofinet,qin2022geometric}
on RTE, RRE and SR. As shown in Table \ref{table:kitti} and Figure~\ref{Qualitative4}, our method achieves the 
highest SR and lowest RTE on KITTI, which means that our method is 
also competitive on outdoor datasets.

\subsection{3DMatch-noise: Robustness to Strong Noise}\label{3DMatch-noise}
From the registration results, 
it can be seen that the most recent methods have determined 
good effects on the 3DMatch dataset, 
and the improvement is close to saturation. 
Therefore, a more challenging benchmark needs to be set.
By adding high-intensity noises to the 3DMatch dataset, 
we construct a 3DMatch-noise benchmark  
to simulate the uncertainty of the real scenes and the imprecision of the sensors.
Our high-intensity noises can be divided into four categories. 
The first category is a Gaussian noise $N(0,0.05^2)$ (clipped to [-0.05, 0.05]), 
which is shown in Figure \ref{noisedata} (b).
The second category shown in Figure \ref{noisedata} (c) is a noise subject to 
the uniform distribution $U(-0.05,0.05)$. 
As shown in Figure \ref{noisedata} (d), the third category 
is to randomly replace 5\% of the original points 
with random numbers that obey the normal distribution $N(0,0.5^2)$. For the fourth category,
considering that 3DMatch is collected by using an RGB-D sensor, 
so it makes more sense to introduce noise from the depth map~\cite{mallick2014characterizations}.
On the original depth map
of 3DMatch, a residual that obeys Gaussian distribution $N(0,0.05^2)$ is added
to the depth channel of the depth map to generate a point cloud with 
Noise 4~\cite{mallick2014characterizations}.

We experiment on the 3DMatch-noise benchmark to further 
highlight the anti-noise ability of our SphereNet in the face of scenes with high-intensity noise. 
As shown in the Table~\ref{noise}, compared with the state-of-the-arts \cite{huang2021predator,yu2021cofinet ,qin2022geometric,li2022lepard},
in the scenes with strong noise, our method has a huge improvement over IR, FMR, and RR. 
Although our SphereNet doesn't use noise augmentation such as \cite{huang2021predator,yu2021cofinet ,qin2022geometric,li2022lepard},
it still achieves the best performances in four different types of noises,
improving RR by 23.2 pp and FMR by 30.5 pp over the state-of-the-art method \cite{qin2022geometric} on Noise 1.
Moreover, after adding Noise 1, Noise 2, Noise 3, Noise 4 to 3DMatch, the RR obtained by our SphereNet only drops by 8.5 pp, 11.8 pp, 1.1 pp
and 5.9 pp,
while 38.3 pp, 31.0 pp, 5.8 pp and 26.9 pp for~\cite{qin2022geometric}.
This highlights that our SphereNet has good robustness to strong noise 
and it also maintains accurate registration even in bad scenes with noise.
As the qualitative results are shown in Figure \ref{qr}, our SphereNet achieves correct registration with the low RRE and RTE.

To verify the stability and influence
under different noises with different intensities, 
we set more abundant experiments on the 3DMatch 
dataset with noise. It can be seen from Table~\ref{table:Different Noise1} that, 
within a certain range, our method has a certain adaptability 
to the noise of different intensities. With the increase in the proportion of Noise 3, 
FMR only decreases in a small range. With the proportion of noise up to 9\%, our RR score is still 85.6\%.
In contrast, our sphere is more sensitive to Noise 1, 
and RR score is only 51.8 when STD is 0.07. But it is still much higher than other algorithms.
\begin{table}[!t]
   \caption{
      Detailed results with different noise, where Noise Std is the standard deviation
      of Noise 1 and proportion is the ratio of the added noise points to the original point cloud in Noise 3.
      }
   \scriptsize
   \setlength{\tabcolsep}{4pt}
   \centering
   \begin{tabular}{c|ccc|c|ccc}
   \toprule
   \multicolumn{4}{c|}{Noise 1} &\multicolumn{4}{c}{Noise 3}  \\
   Noise Std& FMR & IR & RR &Proportion& FMR & IR & RR \\
   \midrule
   0.00   &\textbf{98.2} & \textbf{52.6} & \textbf{91.2}  &0\%   &\textbf{98.2} & \textbf{52.6} & \textbf{91.2}\\
   0.01   &95.8&43.4&86.7  &3\%  &97.2&44.9&88.2\\
   0.03  &89.6&38.0&70.2 &5\%  &{96.7}&{43.2}&88.2\\
   0.05  &{85.5}  & {32.2} & {66.4}  &7\%   &96.1&41.9&87.6\\
   0.07   &72.9&21.7&51.8 &9\%   &95.9&41.0&85.6\\
   \bottomrule
   \end{tabular}
   \label{table:Different Noise1}
\end{table}

\begin{figure*}[!t]
   \centering{\includegraphics[scale=0.85]{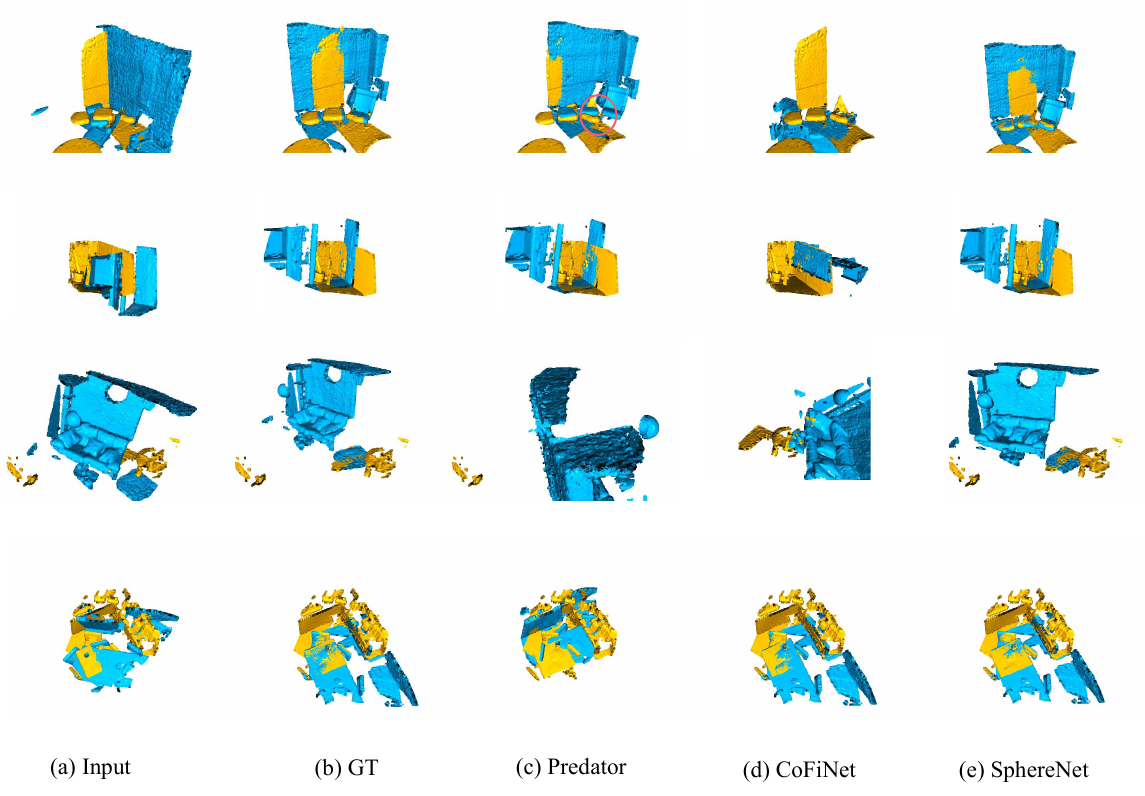}}  
   \caption{Registration results on the 3DMatch and 3DLoMatch datasets.
   }
   \label{Qualitative1}
\end{figure*}


\begin{figure*}[!t]
   \centering{\includegraphics[scale=0.85]{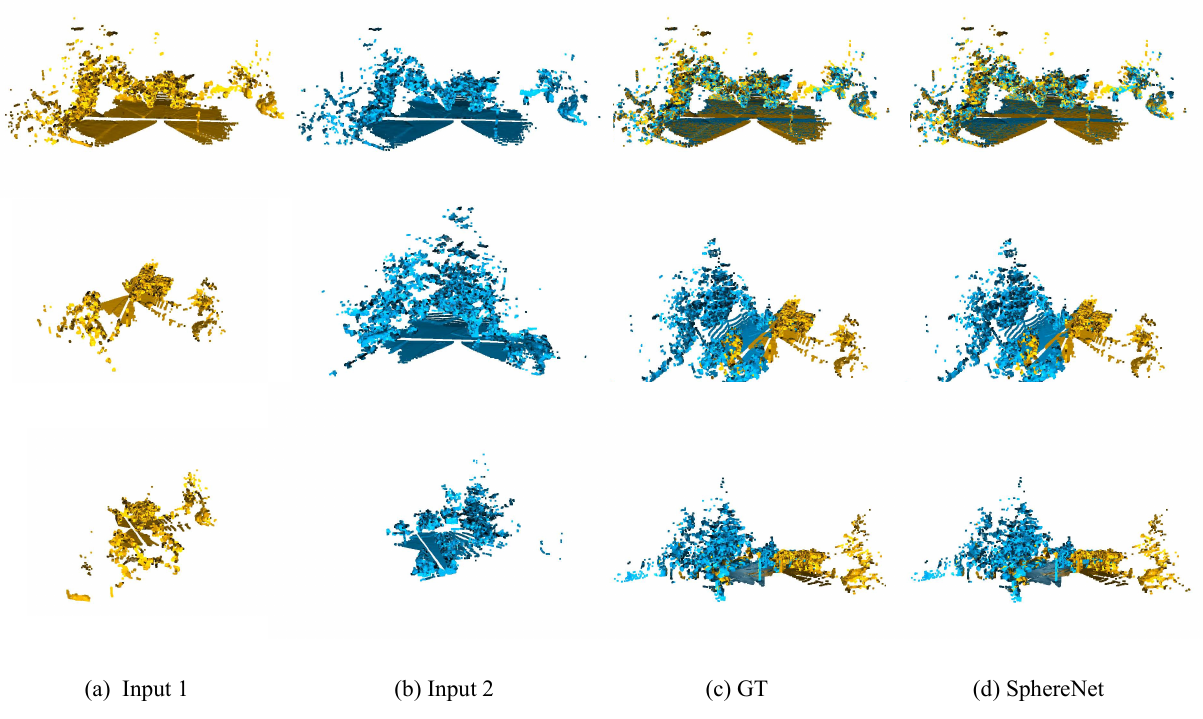}}  
   \caption{
      Registration results on the ETH dataset.
   }
   \label{Qualitative3}
\end{figure*}

\begin{figure*}[!t]
   \centering{\includegraphics[scale=0.9]{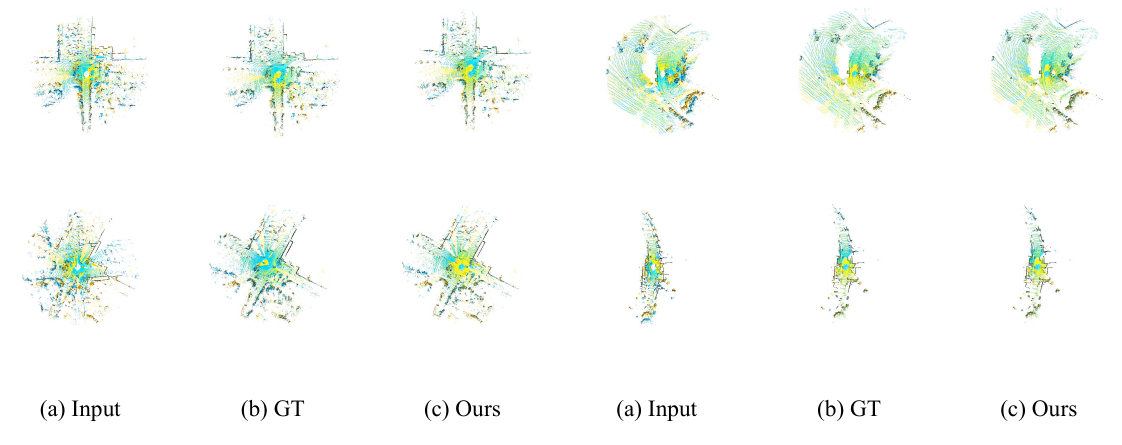}}  
   \caption{
      Registration results on the KITTI dataset.
   }
   \label{Qualitative4}
\end{figure*}


\subsection{Generalization across Different Datasets}

Following \cite{ao2021spinnet,bai2020d3feat}, we also evaluate the 
generalization ability of our SphereNet on the unseen datasets. 
In each experiment, our model is trained on one dataset, 
and multiscale  feature fusion is used to overcome the differences 
in the scale between different datasets. Then, the trained model 
is directly tested on another dataset.
\begin{table}[htbp]
   \caption{Results of generalization from 3DMatch to ETH.}
   \scriptsize
   \centering
			\begin{tabular}{l|cc|cc|c}
            \toprule
				\multirow{2}{*}{}  &
				\multicolumn{2}{c|}{\bf{Gazebo}}   & \multicolumn{2}{c|}{\bf{Wood}}     & \multirow{2}{*}{Avg.}\\
				& Summer        & Winter        & Autumn        & Summer        &                          \\ 
            \midrule
				FCGF~\cite{choy2019fully}   & 22.8          & 10.0          & 14.8          & 16.8          & 16.1                     \\
				D3Feat (rand)~\cite{bai2020d3feat}   & 45.7          & 23.9          & 13.0          & 22.4          & 26.2                     \\
				LMVD~\cite{li2020end}  & 85.3          & 72.0          & {84.0}          & {78.3}          & {79.9}                     \\			
				SpinNet~\cite{ao2021spinnet}   & \underline{92.9} & \underline{91.7} & \textbf{92.2} & \textbf{94.4} & \underline{92.8}           \\ 
            GLORN~\cite{sanchez2022domain}   & 90.3 &91.6  &78.5  &83.2  & 85.9           \\ 
            SphereNet (\emph{ours})   & \textbf{94.6} & \textbf{95.8} & \textbf{92.2} & \underline{93.6} & \textbf{94.5}            \\ 
              \bottomrule
			\end{tabular}
	\label{tab:eth}
\end{table}

\ptitle{Generalization from 3DMatch to ETH Dataset.} 
All our models are trained on the 3DMatch dataset 
and then tested on the ETH dataset with multiscale  feature fusion. As shown in 
Table \ref{tab:eth} and Figure~\ref{Qualitative3}, we achieve the best results in 
three different scenes, and the total average FMR reaches 94.53\%
which is close to 2 pp higher than the state-of-the-art \cite{ao2021spinnet}.

\begin{table}[htbp]
   \caption{Results of generalization from KITTI to 3DMatch.}
   \scriptsize
   \centering
	\begin{tabular}{r|cc|cc|c}
      \toprule
		\multirow{2}{*}{} & \multicolumn{2}{c|}{\textbf{RTE (cm)}} & \multicolumn{2}{c|}{\textbf{RRE ($^\circ$)}} & \multirow{2}{*}{\textbf{Success (\%)}} \\
		& AVG                & STD               & AVG               & STD              &                       \\ 
      \midrule
		FCGF~\cite{choy2019fully}       & 27.1               & 5.58              & 1.61              & 1.51             & 24.19                 \\
		D3Feat-rand~\cite{bai2020d3feat}       & 37.8                   & 9.98                  & 1.58                  & 1.47                 & 18.47                      \\
		D3Feat-pred~\cite{bai2020d3feat}     & 31.6               & 10.1              & 1.44              & 1.35             & \underline{36.76}                 \\
		SpinNet~\cite{ao2021spinnet}     & 15.6               & {1.89}                  & {0.98}                  & {0.63}                 & {81.44}                      \\ 
      SphereNet (\emph{ours})      & \textbf{9.9}               & \textbf{0.71}                  & \textbf{0.57}                  & \textbf{0.33}                 & \textbf{87.20}                      \\ 		        
      \bottomrule
	\end{tabular}
  \label{tab:3dmatch_kitti}
\end{table}

\ptitle{Generalization from 3DMatch to KITTI Dataset.}
Our models are trained on the 3DMatch dataset
and then directly tested on the KITTI dataset. Similarly, it is 
difficult to generalize from 3DMatch 
to KITTI across different sensors. As shown in Table \ref{tab:3dmatch_kitti}, 
the generalization ability of our method also 
shows the best performance on KITTI, 
achieving the minimum AVG and STD in RTE and RRE. 
And SR of the registration is also up to 87.20\%, 
which is nearly 6 pp higher than the previous best method \cite{ao2021spinnet}.

\subsection{Running Times}\label{AE.4}

\begin{table}[thb]
   \caption{Comparison of the running times.}
   \scriptsize
   \setlength{\tabcolsep}{4pt}
   \centering
   \begin{tabular}{lcccc}
   \toprule
   Method &Samples & $t_1 (s/pc)$ & $t_2 (s/pcp)$  & $t_{total} (s/pcp)$  \\
   \midrule
   PerfectMatch\cite{gojcic2019perfect} & 5000 & 9.15 & \textbf{0.19}&18.49  \\
   SpinNet\cite{ao2021spinnet} & 5000 & 11.76 & 0.62 & 24.14 \\
   SphereNet (\emph{ours}) & 5000 & \textbf{3.90} &0.61  &\textbf{8.41}  \\
   \toprule
   \end{tabular}
   \label{tab:running times}
\end{table}

Previous works \cite{wang2021you,yu2021cofinet,ao2022YOTO} have shown that 
the speed of patch-based methods is much slower than that of fragment-based methods. 
Although our SphereNet has some disadvantages in running time, 
it has greatly improved in speed compared with previous patch-based methods \cite{ao2021spinnet,gojcic2019perfect}. 
The running times of different stages are listed in Table~\ref{tab:running times}. 
As shown in Table~\ref{tab:running times}, compared with the previous patch-based methods, 
our method has a significant improvement. 
For completing the registration of one point cloud pair, 
the speed of our Sphere has improved by approximately  65\% compared to SpinNet \cite{ao2021spinnet}.

\subsection{Ablation Study}

In order to better evaluate the effectiveness 
of each module in our SphereNet, we conduct ablation experiments 
on 3DMatch and 3DMatch-noise. In the ablation experiment, 
we used three metrics FMR, IR, and RR to evaluate our results.


\begin{table*}[t]
   \caption{Analysis experiments on 3DMatch / 3DMatch-noise / ETH. 
   \textbf{N, M, K}: Number of spherical voxels in three dimensions. 
   \textbf{R}: Radius of the descriptor. 
   \textbf{SV}: Our spherical voxelization. 
   \textbf{SG}: Spherical voxelization by sphere grouping. 
   \textbf{TI}: Spherical interpolation. 
   \textbf{MSF}: Multiscale feature fusion. 
   \textbf{SP}: Spherical integrity padding. 
   \textbf{ZP}: Zero padding.
   \textbf{MLP}: Use MLP for feature extraction. 
   \textbf{SCNN}: Use SCNN for feature extraction.}
   \resizebox{\linewidth}{!}{
   \begin{tabular}{c|cccccccccccc|cc|cc|cc}
   \hline
   
   \multirow{2}{*}{No.}                & \multirow{2}{*}{N}          & \multirow{2}{*}{M}          & \multirow{2}{*}{K}          & \multirow{2}{*}{R}          & \multirow{2}{*}{SV}         & \multirow{2}{*}{SG}         & \multirow{2}{*}{TI}         & \multirow{2}{*}{MSF}        & \multirow{2}{*}{SP}         & \multirow{2}{*}{ZP}         & \multirow{2}{*}{MLP}        & \multirow{2}{*}{SCNN}  & \multicolumn{2}{c|}{3DMatch}                      & \multicolumn{2}{c|}{3DMatch-noise}           & \multicolumn{1}{c}{ETH}                     \\ 
                                       &&&&&&&&&&&&& FMR(\%) &RR(\%) & FMR(\%) &RR(\%) & FMR(\%)  \\
                                       \hline
                           1)          &15           &20          &40          &0.3         &\checkmark  &            &\checkmark  &            &\checkmark  &            &            & \checkmark &\textbf{98.2}                 &\textbf{91.2}                &\underline{77.7}                &53.8               &\underline{94.5}                             \\
                           2)          &15           &20          &40          &0.5         &\checkmark  &            &\checkmark  &            &\checkmark  &            &            & \checkmark &\underline{98.1}                 &\underline{90.8}                &76.1                &\underline{55.4}               &90.5                              \\
                           3)          &15           &20          &40          &1.0         &\checkmark  &            &\checkmark  &            &\checkmark  &            &            & \checkmark &98.0                 &90.2                &74.6                &55.1               &41.6                              \\
                           4)          &15           &40          &80          &0.3         &\checkmark  &            &\checkmark  &            &\checkmark  &            &            & \checkmark &98.0                 &90.1               &76.1                &\underline{55.4}               & 74.1                            \\
                           5)          &15           &20          &40          &0.3         &            &\checkmark  &            &            &\checkmark  &            &            & \checkmark &75.5                 &52.5                &32.1                &14.1               & 36.3                             \\
                           6)          &15           &20          &40          &0.3         &\checkmark  &            &            &            &\checkmark  &            &            & \checkmark &70.6                 &46.2                &13.3                &4.3               &  16.7                            \\
                           7)          &15           &20          &40          &0.3         &\checkmark  &            &\checkmark  &\checkmark  &\checkmark  &            &            & \checkmark & \underline{98.1}               &90.5                &\textbf{85.5}                 &\textbf{66.4}              &\textbf{95.8}                               \\
                           8)          &15           &20          &40          &0.3         &\checkmark  &            &\checkmark  &            &            &\checkmark  &            & \checkmark & 95.2               &84.9                &72.6                &52.5               & 74.3                             \\
                           9)          &15           &20          &40          &0.3         &\checkmark  &            &\checkmark  &            &\checkmark  &            & \checkmark &            &51.3                 &31.6                &10.3                &9.6               & 9.5                             \\
                           \hline
   \end{tabular}}
   
   \label{tab:ablation1}
   \end{table*}

\ptitle{(1) Replacing the spherical voxelization with the sphere grouping.}
Spherical voxelization can well capture 
the geometric feature from the 
perspective of geometric structure, and it can 
cooperate well with our spherical CNN. In this
experiment, we replacing the spherical voxelization 
with the sphere grouping to form the initial feature 
by searching the nearest neighbors in the 
fixed position.
By comparing the Row 1 and 5 in Table~\ref{tab:ablation1},
without using the spherical voxelization to encode the initial feature, 
the ablated method has a lower FMR score than the complete method. And also,
the performances on 3DMatch-noise and ETH significantly reduce. 
This illustrates that spherical voxelization is a great contribution to the finer 
geometric feature and better generalization ability.

\ptitle{(2) Only removing the spherical interpolation.}
In order to ensure the fineness of the initial 
feature with less voxel grids, 
we introduce the spherical interpolation, 
which can improve the robustness against noise. 
In this experiment, we remove the spherical interpolation 
and directly use the initial feature generated by 
spherical voxelization to extract the descriptor.
As shown in Row 1 and 6, without using the spherical interpolation, 
the ablated method performs poorly on the 3DMatch-Noise benchmark and ETH dataset.
This shows that our spherical interpolation 
can greatly improve the anti-noise ability and generalization ability.

\ptitle{(3) Only replacing the spherical integrity padding with zero padding.}
Due to the particularity of our spherical voxelization, 
the obtained initial feature has some certain loss on the 
boundaries, and our spherical integrity padding can solve 
this problem well. In this experiment, we replace the spherical 
integrity padding with zero padding.
By comparing the Row 1 and 8 in Table~\ref{tab:ablation1},
the method with spherical integrity padding has improved RR from 84.9\%
to 91.2\%, which important to improve the performance of our method.
In addition, it also has a contribution to anti-noise ability.

\ptitle{(4) Only replacing the spherical CNN with MLPs.}
Our spherical CNN can cooperate with the spherical voxelization to 
complete the extraction of rotation-invariant and robust 
descriptors. In this experiment, only MLP is used instead of 
spherical CNN to verify the significant improvement 
of our spherical CNN for registration.
As shown in Row 1 and 9, 
without spherical CNN, the ablated method cannot achieve accurate feature matching and registration,
which shows that our spherical CNN are crucial 
for this robust descriptor extraction. 

\ptitle{(5) Only removing the Multiscale Feature Fusion.}
Our MSF module introduces a scale channel  and refines the initial 
features without new overhead to further improve the generalization ability and anti-noise ability. 
As can be seen from Row 1 and 7 of Table~\ref{tab:ablation1}, 
our method with MSF module 
achieves a 1.3\% improvement in RR on the ETH dataset and a 12.6\% 
improvement on the 3DMatch-noise benchmark.

\ptitle{Why does the model have better generalization?}

First, \textbf{MSF module}. Comparing Row 1 and 7 in Table~\ref{tab:ablation1}, 
The generalization performance is improved by using MSF module.
Although the generalization performance of the local context~\cite{ao2021spinnet}
is excellent, we still achieve better results than 
SpinNet~\cite{ao2021spinnet} by our MSF module.
It is difficult in unseen datasets because of the inconsistent scale and sparsity. 
So a suitable radius of the descriptor is necessary  
to focus on the local area,
such as ``a chair" in 3DMatch or ``a vehicle" in KITTI which is more distinctive for matching. 
SpinNet~\cite{ao2021spinnet} only adjusts the radius of the descriptor manually which has a limited effect. 
This paper proposes the MSF module, which divides a local sphere into multiscale spheres, 
then carries out feature fusion. 
In this way, the ``suitable area" can be better captured, which greatly improves the generalization ability.

Second, \textbf{spherical voxelization}. Comparing the results of the first and fifth rows in Table~\ref{tab:ablation1}, 
it can be seen that our spherical interpolation method has greatly contributed to 
the improvement of generalization ability. This is because, 
different from SG method (Row 5) in SpinNet~\cite{ao2021spinnet}, 
the area size of our generated voxels is regular and proportional to the descriptor radius $R$ of our method. 
Therefore, our  SV method can adapt the size of a small voxel according to the descriptor radius $R$, 
which can better improve the adaptability of our method to the scale inconsistency of different datasets.
However, SG method uses the neighbor search with a fixed radius to generate small voxels, 
which can't adapt to the scale inconsistency of different datasets well.

\ptitle{Why does the model have anti-noise ability?}

First, \textbf{spherical voxelization}. As shown in Row 1 and 5 on Table~\ref{tab:ablation1},
Our method on spherical voxelization has a better contribution on 
anti-noise ability than method in SpinNet~\cite{ao2021spinnet}.
Analyzing the cause, SpinNet~\cite{ao2021spinnet} builds the voxels by performing radius neighbor 
search at fixed locations and setting  the number 
of voxel grids to 9*40*80, which leads to a high overlap of the 
regions where point-based layers are performed. 
So a small amount of noise will affect more histogram bins.
In our method, the sphere is strictly divided into voxels in three directions, 
and the number of voxels is fewer. There is no overlapping area 
in the histogram generation. So the noise will only affect fewer histogram bins. 

Second, \textbf{spherical interpolation}. Comparing Row 1 and 6 in Table~\ref{tab:ablation1}, 
using TI achieves a great improvement on RR and without it RR is only 4.3\%. Therefore,
our TI has a good contribution to the improvement of anti-noise ability.
As we describe in Section \ref{Spheroid}, a point is interpolated to the 8 surrounding voxels, rather than voting completely to a voxel.
many points affected by noise will be offset from the position $\mathbf{p}_{i_j} \in \mathbf{V}_{n_1 m_1 k_1}$
to  $\mathbf{p}_{i_j}+\boldsymbol\varepsilon \in \mathbf{V}_{n_2 m_2 k_2}$. 
When the noise intensity is within a certain range, both $\mathbf{V}_{n_1 m_1 k_1}$ and $\mathbf{V}_{n_2 m_2 k_2}$
belong to the 8 surrounding voxels. 
Therefore, our spherical interpolation can weaken the effect of noise on the initial features.

\section{Conclusion}
We propose SphereNet to learn a noise-robust and 
unseen-general descriptor for point cloud registration.  
We encode the initial geometric feature by spherical voxelization with spherical interpolation
so that we can fully capture the geometric features. 
After that, descriptors are extracted by spherical CNN with spherical integrity padding, 
which makes the descriptors more robust and general. 
Thanks to these methods, SphereNet achieves the state-of-the-art 
performance compared with other learning-based methods.
In addition,  it also achieves the best anti-noise ability on the noisy scenes and generalization performance
on the unseen scenes.
In the future, by the research of new technologies, we
will use SphereNet framework to complete the task of non-rigid point cloud registration.

{\small
\bibliographystyle{ieee_fullname}
\bibliography{egbib}
}

\end{document}